%% file: neurips.tex
\documentclass{article}

\usepackage[preprint, nonatbib]{neurips_2024}
\usepackage[utf8]{inputenc} % allow utf-8 input
\usepackage[T1]{fontenc}    % use 8-bit T1 fonts
\usepackage{hyperref}       % hyperlinks
\usepackage{url}            % simple URL typesetting
\usepackage{booktabs}       % professional-quality tables
\usepackage{amsfonts}       % blackboard math symbols
\usepackage{nicefrac}       % compact symbols for 1/2, etc.
\usepackage{microtype}      % microtypography
\usepackage{xcolor}         % colors
\usepackage{amsthm}
\usepackage{thmtools}
\usepackage{bbm}
\usepackage{mathtools}
\usepackage{amsmath}
\usepackage{amsthm}
\usepackage{amssymb}
\usepackage{mathtools}
\usepackage{enumitem}
\usepackage{wrapfig}
\usepackage{dsfont}
\usepackage[ruled, noend]{algorithm2e}
\usepackage{multicol}
\usepackage{multirow}
\usepackage{hhline}
\usepackage{xcolor,colortbl}
\usepackage{caption}
\usepackage{subcaption}
\usepackage{wrapfig}
\usepackage{enumitem,kantlipsum}
\usepackage{amssymb}
\usepackage{threeparttable}
\usepackage{lipsum} % For filler text
\usepackage{minitoc}
\usepackage{tabularx}
\usepackage{graphicx}
\usepackage{calligra}
\usepackage{algorithmic}
\usepackage{minitoc}

\definecolor{dwhred}{rgb}{0.44, 0.72, 0.93}
\definecolor{dwhblue}{rgb}{0.96, 0.49, 0.43}
\definecolor{dwhgreen}{rgb}{0.33, 0.82, 0.47}

\input{math_commands.tex}

\theoremstyle{plain}
\newtheorem{theorem}{Theorem}

\theoremstyle{definition}
\newtheorem{definition}{Definition}
\newtheorem{assumption}{Assumption}
\theoremstyle{remark}

\newcommand{\method}{OASIS}
\newcommand{\methodlong}{OASIS (c\textbf{O}ndition\textbf{A}l di\textbf{S}tribut\textbf{I}on \textbf{S}haping)}

\title{OASIS: Conditional Distribution Shaping for \\ Offline Safe Reinforcement Learning}

\author{%
  Yihang Yao$^*$$^1$, Zhepeng Cen$^*$$^1$, Wenhao Ding$^{1}$, Haohong Lin$^{1}$, Shiqi Liu$^1$, \\
  \textbf{Tingnan Zhang$^2$, Wenhao Yu$^2$, Ding Zhao$^1$} \\
  $^1$ Carnegie Mellon University, $^2$ Google DeepMind\\ 
  $^*$ Equal contribution, \texttt{\{yihangya, zcen\}@andrew.cmu.edu}\\
}

\begin{document}
\doparttoc % Tell to minitoc to generate a toc for the parts
\faketableofcontents % Run a fake tableofcontents command for the partocs

\maketitle

\begin{abstract}
Offline safe reinforcement learning (RL) aims to train a policy that satisfies constraints using a pre-collected dataset. Most current methods struggle with the mismatch between imperfect demonstrations and the desired safe and rewarding performance. In this paper, we introduce \methodlong, a new paradigm in offline safe RL designed to overcome these critical limitations. \method\ utilizes a conditional diffusion model to synthesize offline datasets, thus shaping the data distribution toward a beneficial target domain. Our approach makes compliance with safety constraints through effective data utilization and regularization techniques to benefit offline safe RL training. Comprehensive evaluations on public benchmarks and varying datasets showcase \method's superiority in benefiting offline safe RL agents to achieve high-reward behavior while satisfying the safety constraints, outperforming established baselines. Furthermore, \method\ exhibits high data efficiency and robustness, making it suitable for real-world applications, particularly in tasks where safety is imperative and high-quality demonstrations are scarce.
\end{abstract}

\vspace{-5pt}
\section{Introduction}
\vspace{-5pt}
% Offline safe RL motivation: safety constraints in offline RL and decision-making
Offline Reinforcement Learning (RL), which aims to learn high-reward behaviors from a pre-collected dataset \cite{prudencio2023survey, fu2020d4rl}, has emerged as a powerful paradigm for handling sequential decision-making tasks such as autonomous driving \cite{stachowicz2024racer, lin2024safety, zhang2023spatial, fang2022offline}, robotics \cite{chi2023diffusion, ding2024seeing, zhao2023guard, li2023guided}, and control systems \cite{zhan2022deepthermal}. Although standard offline RL has achieved remarkable success in some environments, many real-world tasks cannot be adequately addressed by simply maximizing a scalar reward function due to the existence of various safety constraints that limit feasible solutions. The requirement for \textit{safety}, or constraint satisfaction, is particularly crucial in RL algorithms when deployed in real-world tasks~\cite{kim2022safety, he2024agile, xiao2023safe, huang2023safedreamer,chen2023progressive, zou2024policy, yang2023model}.

% Introduce offline safe RL
To develop an optimal policy within a constrained manifold \cite{garcia2015comprehensive, brunke2021safe}, \textit{offline safe RL} has been actively studied in recent years, offering novel ways to integrate safety requirements into offline RL \cite{wachi2024survey}. Existing research in this area incorporates techniques from both offline RL and safe RL, including the use of stationary distribution correction \cite{lee2022coptidice, lee2021optidice}, regularization \cite{kostrikov2021offline, xu2022constraints}, and constrained optimization formulations \cite{le2019batch}. Researchers have also proposed the use of sequential modeling methods, such as the decision transformer \cite{liu2023constrained} and the decision diffuser \cite{lin2023safe, zheng2024safe} to achieve advantageous policies and meet safety requirements.

% Limitation of existing offline safe RL
% their application is limited by several challenges. First, 
Although these methods show promise, it is difficult to handle state-action pairs that are absent from the dataset, which is known notably as out-of-distribution (OOD) extrapolation issues~\cite{zheng2024safe, fujimoto2019off, shi2022pessimistic}.
To solve this, many works utilize regularization methods to push the policy toward behavior policy to achieve pessimism~\cite{fujimoto2019off,shi2022pessimistic}. However, this approach worsens the situation when the dataset is imbalanced and biased: regularization by imperfect demonstrations such as datasets composed predominantly of low-reward data or containing few safe trajectories collected using unsafe behavior policies. This regularization also leads to another challenge: striking the optimal balance between learning objectives such as task utility efficiency and safety requirements, leading to reward degradation or aggressive behavior~\cite{liu2023constrained, yao2023gradient, hong2023beyond}.
% The low quality of the safe dataset further makes the offline safe RL problem hard to solve.

\begin{wrapfigure}{R}{0.44\textwidth}
\vspace{-9mm}
\centering
\includegraphics[width=1.0\linewidth]{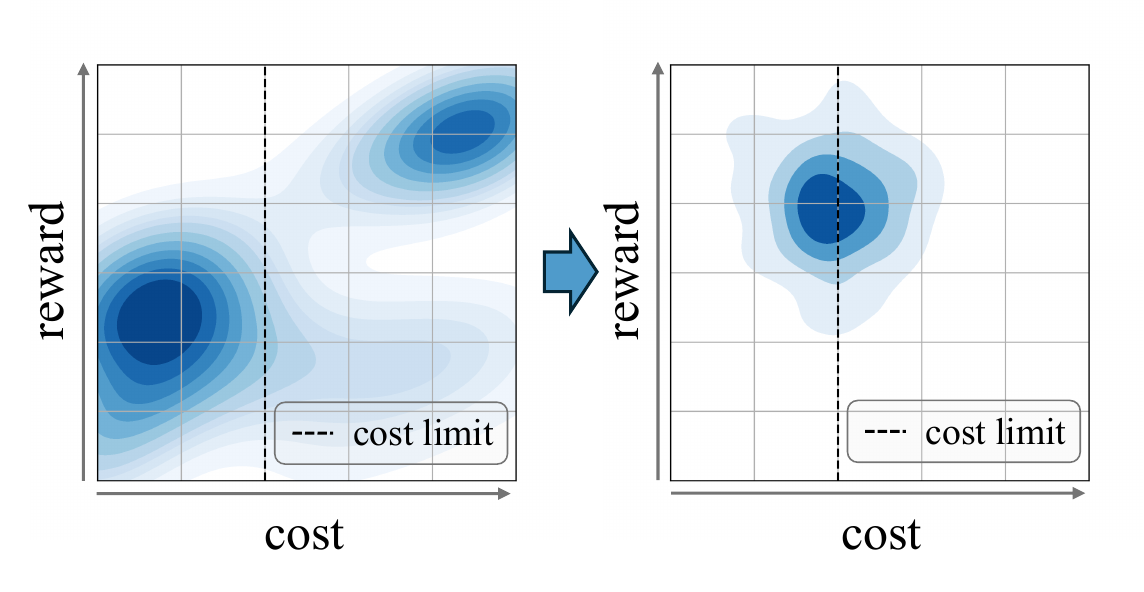}
\vspace{-7mm}
\caption{An example of distribution shaping in offline safe RL. We generate a low-cost and high-reward dataset from the original dataset for subsequent RL training.
}
\label{Fig: distribution shaping}
\vspace{-5mm}
\end{wrapfigure}

% Contribution Summary
To address these challenges, we introduce a new paradigm in offline safe RL, \methodlong, which steers the offline data distribution to a beneficial target domain using a conditional diffusion model as shown in Figure~\ref{Fig: distribution shaping}. \method\ distills knowledge from the imperfect dataset, and generates rewarding and safe data according to the cost limit to benefit offline safe RL training. This approach is designed to be compatible with general Q-learning-based offline safe RL algorithms.
The key contributions are summarized as follows.

\noindent \textbf{1. Identification of the safe dataset mismatch (SDM) problem in offline safe RL.}  We identify the mismatch between the behavior policy and the target policy and investigate the underlying reasons for performance degradation with this condition. 
% This insight spurred the development of \method, which employs data distribution shaping to address this issue.

% This perspective highlights the limitations of most existing constrained-optimization-based approaches and motivates the development of CCPO based on conditional variational inference. Importantly, CCPO can generalize to diverse unseen constraint thresholds without retraining the policy.

\noindent \textbf{2. Introduction of the \method\ method to address the SDM problem.} 
To the best of our knowledge, this is the first successful application of a distribution shaping paradigm within offline safe RL. Our theoretical analysis further provides insights into performance improvement and safety guarantees.

\noindent \textbf{3. A comprehensive evaluation of our method across various offline safe RL tasks.} The experiment results demonstrate that \method\ outperforms baseline methods in terms of both safety and task efficiency for varying tasks and datasets. 
% In addition, our method reduces computation requirements by improving data efficiency and shows a consistent superior performance with different hyperparameters.

% \methodlong

% Potential questions to answer (or leave them to rebuttal):
% \begin{enumerate}
%     \item The empirical performances on some tasks are not good enough (safety-gymnasium navigation tasks).
%     \item the authors propose to use generative model to augment data under the certain condition. Are the other generative models able to achieve similar performances?
%     \item theoretical analysis, connect it to diffusion model, and why can the proposed method reduce the gap of data distribution and optimal distribution
%     \item More ablation studies on hyper-parameters (condition, other hyperparameters in diffusion model).
% \end{enumerate}
% \input{tex/introduction}

\vspace{-5pt}
\section{Related Work}
\vspace{-5pt}
\textbf{Offline RL.} Offline RL addresses the limitations of traditional RL, which requires interaction with the environment. The key literature on offline RL includes BCQ~\cite{fujimoto2019off}, which mitigates the extrapolation error using a generative model, and CQL~\cite{kumar2020conservative}, which penalizes the overestimation of Q-values for unseen actions. BEAR~\cite{kumar2019stabilizing} further addresses the extrapolation error by constraining the learned policy to stay close to the behavior policy. OptiDICE~\cite{lee2021optidice} directly estimates the stationary distribution corrections of the optimal policy, and COptiDICE~\cite{lee2022coptidice} extends the method to the constrained RL setting.
Recent advances have increasingly focused on the use of data generation to improve policy learning. S4RL~\cite{sinha2022s4rl} shows that surprisingly simple augmentations can dramatically improve policy performance. \cite{yu2022leverage} explores leveraging unlabeled data to improve policy robustness, while \cite{li2024survival} proposes survival instincts to enhance agent performance in challenging environments.
% COMBO~\cite{yu2021combo} integrates conservative principles into model-based policy optimization to improve stability. 
Counterfactual data augmentation is another promising direction in offline RL~\cite{pitis2020counterfactual, aloui2023counterfactual, lu2020sample, pitis2022mocoda}, highlighting the potential of data generation to significantly improve efficiency and effectiveness.

% Counterfactual data augmentation is another promising direction, with several approaches such as the locally factored dynamics method~\cite{pitis2020counterfactual}, contrastive learning~\cite{aloui2023counterfactual}, Generative Adversarial Networks~\cite{lu2020sample}, and graph-based generation~\cite{pitis2022mocoda}. These innovations highlight the potential of data generation to significantly improve the effectiveness and efficiency of offline RL, thereby motivating this research.

\textbf{Safe RL.} Safe RL is formulated as a constrained optimization to maximize reward performance while satisfying the pre-defined safety constraint~\cite{garcia2015comprehensive, achiam2017constrained, gu2022review, liu2022robustness, kim2024trust, xu2023uncertainty, kim2024scale}. Primal-dual framework is one common approach to solve safe RL problem~\cite{chow2018risk, tessler2018reward, ray2019benchmarking, ding2020natural, zhang2020first, cen2024feasibility, wu2024off, ding2024resilient}. To mitigate the instability issue during online training, \cite{stooke2020responsive} proposes PID-based Lagrangian update while \cite{liu2022constrained} and \cite{huang2022constrained} apply variational inference to compute optimal multiplier directly. Another line of work for safe RL is to extend to offline settings, which learn from a fixed dataset to achieve both high reward and constraint satisfaction~\cite{guan2024poce, zhang2023saformer}. \cite{xu2022constraints, polosky2022constrained, hong2024primal, guan2024voce} tailor online prime-dual-style algorithms to reduce the out-of-distribution issue in offline setting. \cite{liu2023constrained, guo2024temporal} use decision transformer~\cite{chen2021decision} to avoid the value estimation and exhibit consistent performances across various tasks.

% Regularization: Staying in the data support~\cite{li2024survival} 
% what is this paper?

% Diffusion-based safe RL~\cite{zheng2024safe, lin2023safe, romer2024safe}
\textbf{Diffusion Models for RL.} Diffusion models have recently gained attention in RL for their capabilities in planning and data generation~\cite{zhu2023diffusion, lee2024gta, yuan2024reward}. Specifically, Diffuser~\cite{janner2022planning} uses a diffusion process to plan the entire trajectory in complex environments. \cite{ajay2022conditional} extends this with the Decision Diffuser, which conditions the diffusion process on specific goals and rewards to improve decision-making. SafeDiffuser~\cite{xiao2023safediffuser} and FISOR~\cite{zheng2024safe} enhance safety by ensuring the planned trajectories satisfying constraints. Combined with the data augmentation capability of diffusion models, AdaptDiffuser~\cite{liang2023adaptdiffuser} achieves state-of-the-art results on offline RL benchmarks. \cite{lu2024synthetic} proposes Synthetic Experience Replay, leveraging diffusion models to create synthetic experiences for more efficient learning. \cite{he2024diffusion} demonstrates that diffusion models are effective planners and data synthesizers for multi-task RL, showcasing their versatility and efficiency. In this work, we investigate the power of diffusion models for safe RL, where the balance between reward and cost presents further complexities.

% \paragraph{Counterfactual Data Augmentation in RL.}
% % vanilla augmentation
% One way to simulate multiple environments is data augmentation. However, most data augmentation works~\cite{laskin2020curl, wang2020improving, yarats2021mastering, kostrikov2020image, hansen2021stabilizing, raileanu2021automatic, hansen2021generalization} apply image transformation to raw inputs, which requires strong domain knowledge for image manipulation and cannot be applied to other types of inputs.
% In RL, the dynamic model and reward model follow certain causal structures, which allow counterfactual generation of new transitions based on the collected samples. 
% This line of work, named counterfactual data augmentation, is very close to this work.
% Deep generative models~\cite{lu2020sample} and adversarial examples~\cite{agarwal2023synthesizing} are considered for the generation to improve sample efficiency in model-based RL.
% CoDA~\cite{pitis2020counterfactual} and MocoDA~\cite{pitis2022mocoda} leverage the concept of locally factored dynamics to randomly stitch components from different trajectories. However, the assumption of local causality may be limited.

% \input{tex/related_work}

\vspace{-5pt}
\section{Problem Formulation}
\vspace{-5pt}
\subsection{Safe RL with Constrained Markov Decision Process}
\label{section: problem formulation}

We formulate Safe RL problems under the Constrained Markov Decision Process (CMDP) framework~\cite{altman1998constrained}. A CMDP $\Mcal$ is defined by the tuple $(\Scal, \Acal, \Pcal, r, c, \gamma, \mu_0)$, where $\Scal \in \mathbb{R}^m$ is the state space, $\Acal\in \mathbb{R}^n$ is the action space, $\Pcal:\Scal \times \Acal \times \Scal \xrightarrow{} [0, 1]$ is the transition function, $r:\Scal \times \Acal \times \Scal \xrightarrow{} \mathbb{R}$ is the reward function, $c:\Scal \times \Acal \times \Scal \xrightarrow{} \mathbb{R}_{\geq 0}$ is the cost function, $\gamma$ is the discount factor, and $\mu_0: \Scal \xrightarrow[]{} [0,1]$ is the initial state distribution. Note that this work can also be applied to multiple-constraint tasks, but we use a single-constraint setting for easy demonstration. A safe RL problem is specified by a CMDP and a constraint threshold $\kappa \in [0, +\infty)$. Denote $\pi \in \Pi:\Scal \times \Acal\rightarrow [0,1]$ as the policy and $\tau = \{(s_1, a_1, r_1, c_1), (s_2, a_2, r_2, c_2), \dots \}$ as the trajectory. The stationary state-action distribution under the policy $\pi$ is defined as $d^\pi(s,a) = (1-\gamma) \sum_{t} \gamma^t \Pr(s_t=s,a_t=a)$. The reward and cost returns are defined as $R(\tau) = \sum_{\tau} r$, and $C(\tau)=\sum_{\tau} c$. 
% The reward return and cost return of a trajectory at timestep $t$ are denoted as $V_r^\tau(s_t) = \sum_{t'=t}^T r_{t'}$ and $V_c^\tau(s_t) = \sum_{t'=t}^T c_{t'}$, respectively.
The value function is $V_\fs^\pi(\mu_0) = \mathbb{E}_{\tau \sim \pi, s_0 \sim \mu_0}[ \sum_{t=0}^\infty \gamma^t \fs_t ], \fs\in \{r, c\}$, which is the expectation of discounted return under the policy $\pi$ and the initial state distribution $\mu_0$. 
The goal of safe RL is to find the optimal policy $\pi^*$ that maximizes the expectation of reward return while constraining the expectation of cost return to the threshold $\kappa$:
\begin{equation}
% \vspace{-2mm}
   \pi^* = \arg\max_{\pi} \mathbb{E}_{\tau \sim \pi}  \big[R(\tau)], \quad s.t. \quad \mathbb{E}_{\tau \sim \pi} \big[C(\tau)] \leq \kappa. 
   \label{eq:safe-rl}
\end{equation} 

\subsection{Regularized offline safe RL}
\label{subsection: regularized offline safe rl}

For an offline safe RL problem, the agent can only access a pre-collected dataset $\Dcal = \cup_i \Dcal_i$, where $\Dcal_i \sim \pi_i^B$ is collected by the behavior policy $\pi_i^B \in \Pi^B$.
% In addition to the challenges such as the Out-of-Distribution (OOD) action issue~\cite{fujimoto2019off, kumar2019stabilizing} brought by offline RL, offline safe RL algorithms also face the balance problem between reward and safety performance.
To solve the problem in Eq.(\ref{eq:safe-rl}), we convert the constraint optimization problem into an unconstrained form:
\begin{equation}
\left(\pi^*, {\lambda}^*\right)=\arg \max _\lambda \min _{\pi} \mathcal{J}(\pi, {\lambda}), \quad \mathcal{J}(\pi, {\lambda}) = - \mathbb{E}_{\tau \sim \pi}  R(\tau) + {\lambda} (\mathbb{E}_{\tau \sim \pi}  R(\tau) - {\kappa}).
\label{equ: lagrangian safe RL}
\end{equation}
The primal-dual-based algorithm solves the optimal policy $\pi^*$ and the dual variable $\lambda^*$ by updating $\left(\pi, {\lambda}\right)$ iteratively \cite{stooke2020responsive, lee2022coptidice, hong2024primal}. In offline safe RL tasks, a regularization term is usually introduced to prevent the action OOD issue~\cite{lin2024policy}, that is, the objective is converted to:
\begin{equation}
\left(\pi^*, {\lambda}^*\right)=\arg \max _\lambda \min _{\pi} \mathcal{J}_{\text{off}}(\pi, {\lambda}), \quad \mathcal{J}_{\text{off}}(\pi, {\lambda}) = \mathcal{J}(\pi, {\lambda}) +w L(\pi, \pi^B),
\label{equ: offline safe RL}
\end{equation}
where $w > 0$ is a constant weight, $L(\pi, \pi^B)$ is a regularization term and $\pi^B$ is the empirical behavior policy and can be viewed as a mixture of $\{\pi_i^B\}$.
Practically, regularization is formulated as the MSE regularization~\cite{fujimoto2021minimalist} or the evidence lower bound regularization~\cite{fujimoto2019off, kumar2019stabilizing}. In offline safe RL, there are two main challenges: (1) \textbf{Distribution shift}~\cite{kostrikov2021offline}. The agent has poor generalizability when facing OOD state-action pairs during online evaluation; and (2) \textbf{Efficiency-safety performance balancing}~\cite{liu2023constrained}. The agent tends to be over-conservative or aggressive when taking an over-estimation or an under-estimation of the safety requirements.
% \TODO{reorganize the last sentence}

% We identify that common ways to solve these two challenges conflict with each other. A universal method to mitigate these two issues simultaneously is needed.

\begin{wrapfigure}{R}{0.4\textwidth}
\vspace{-6mm}
\centering
\includegraphics[width=.98\linewidth]{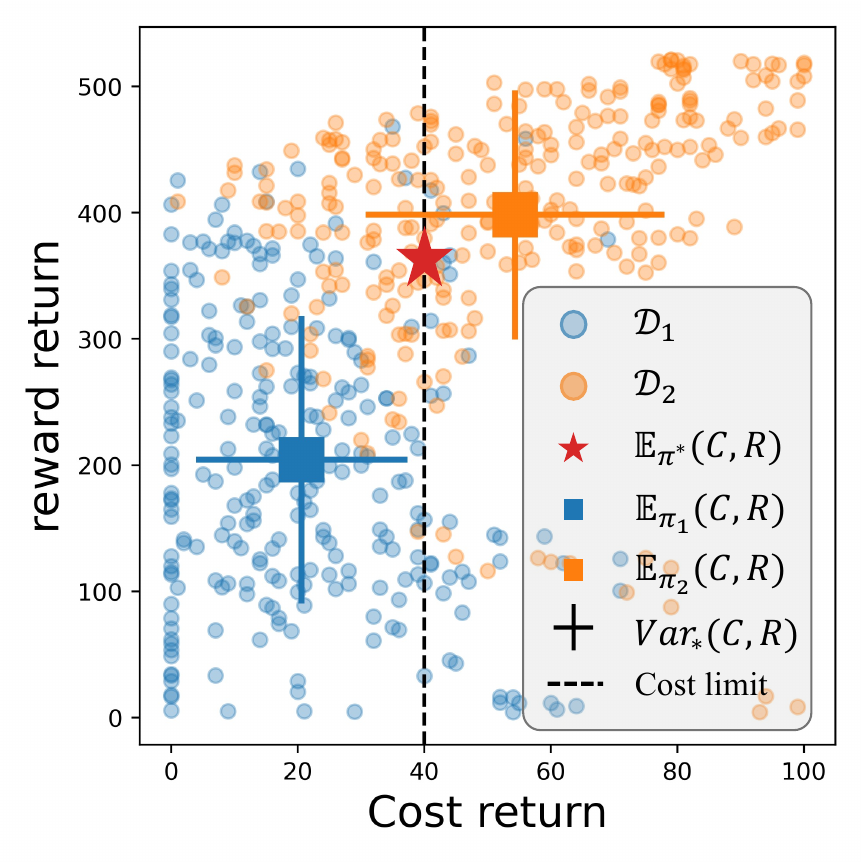}
\vspace{-3mm}
\caption{$\Dcal_1$ is a conservative dataset, and $\Dcal_2$ is a tempting dataset. Each point represents $(C(\tau), R(\tau))$ of a trajectory $\tau$ in the dataset. 
}
\label{Fig: dataset type}
\vspace{-10mm}
\end{wrapfigure}

\vspace{-5pt}
\section{Method}
\vspace{-5pt}
In this section, we first identify the \textit{safe dataset mismatch} (SDM) problem, which leads to performance degradation when solving the regularized offline safe RL objective~(\ref{equ: offline safe RL}). Then we present the proposed \methodlong\ method to solve this problem. The \method\ method utilizes the diffusion model to realize conditional distribution shaping, solving the challenges mentioned above, thus benefiting offline safe RL training. Following the proposed algorithm, we provide a theoretical guarantee of the safety performance of the policy learned in this paradigm.

\subsection{Identification of the Safe Dataset Mismatch Problem}
% \begin{enumerate}
%     \item What is the problem? Tempting problem~\cite{liu2022robustness, liu2023constrained}. Offline RL regularizer~\cite{lin2024policy}. Safe RL as MOO, safety requirements as preference. \cite{lin2024policy, yao2023gradient, zheng2024safe}
%     \item Why is this problem critical and hard for offline safe RL? \cite{liu2023constrained}
% \end{enumerate}

% The SDM problem arises from the imperfect demonstration. 
The regularized offline safe RL objective~(\ref{equ: offline safe RL}) pushes policy to behavior policy to prevent action OOD issues~\cite{shi2022pessimistic}. 
When given an imperfect dataset, the state-action distribution deviates from the optimal distribution, and the SDM problem arises: if the behavior policy is too conservative with low costs and low rewards, it leads to task efficiency degradation; if the behavior policy is too aggressive with high costs and high rewards, it leads to safety violations.
To further investigate the SDM problem and the effect of dataset distribution on offline safe RL, we define the \textbf{tempting dataset} and \textbf{conservative dataset}, which are based on tempting and conservative policies:

\begin{definition}[Tempting policy~\cite{liu2022robustness} and conservative policy]
\label{definition: tempting policy}
The tempting policy class is defined as the set of policies that have a higher reward return expectation than the optimal policy, and the conservative policy class is defined as the set of policies that have lower reward and cost return expectations than the optimal policy: 
% \TODO{$V_r^{\pi}(\mu_0)$ is undefined; use $V_r^{*}(\mu_0)$ to align with the theorem and proof}
\begin{equation}
    \Pi^T \coloneqq \{ \pi: V_r^\pi(\mu_0) > V_r^{\pi^*}(\mu_0)\}, \quad \Pi^C \coloneqq \{ \pi: V_r^\pi(\mu_0) < V_r^{\pi^*}(\mu_0),  V_c^\pi(\mu_0) < V_c^{\pi^*}(\mu_0)\}.
\end{equation}
\end{definition}
Intuitively, a tempting policy is a more rewarding but less safe policy than the optimal one, and a conservative policy is with lower cost but less rewarding.
According to these policies, we define two types of datasets:
\begin{definition}[{Tempting} and {conservative dataset}]
\label{definition: tempting dataset}
For an offline dataset $D_i \sim \pi_i$, if $\pi_i \in \Pi^B \cap \Pi^T$, then the dataset is tempting; if $\pi_i \in \Pi^B \cap \Pi^C$, then the dataset is conservative.
\end{definition}
% Similar to the definition of the tempting dataset, the conservative dataset is defined as:
% \begin{definition}[Conservative dataset]
% \label{definition: conservative dataset}
% \end{definition}
Staying within the tempting dataset distribution results in tempting (unsafe) behavior, while staying within the conservative dataset distribution causes reward degradation. A theoretical analysis of performance degradation due to the SDM problem is presented in section~\ref{subsection: theoretical analysis}. Figure~\ref{Fig: dataset type} illustrates examples of both conservative and tempting datasets.
% Both tempting and conservative datasets can cause conflicting optimization gradients and have negative impacts on regularized offline safe RL:
% \begin{lemma}[Conflicting optimization gradients]
% For an agent trained on conservative or tempting dataset $\mathcal{D}_i$, there exists a neighbor of the optimal policy $Br(\theta^*)$, when $\theta \in Br(\theta^*)$,
% the regularizer gradients and the gradient of safe RL objective $\mathcal{J}$ in (\ref{equ: lagrangian safe RL}) are contradictory, i.e, 
% \begin{equation}
%     \cos (\nabla_{\theta} \mathcal{J}, \nabla_{\theta} L(\pi, \pi_i^B) < 0,
% \end{equation}
% where $\theta^*$ denotes the parameter for $\pi^*$, and $\cos(\cdot, \cdot)$ is the cosine similarity function.
% \label{lemma: conflict optimization}
% \end{lemma}
% The proof is available in Appendix~\ref{appendix-sec: proof}. 
% The COG problem causes the sub-optimality problem and leads to performance degradation.
It is important to note that tempting and conservative datasets are prevalent in offline safe RL since optimal policies are rarely available for data collection. The SDM problem is a distinct feature of offline safe RL, indicating that training the policy on either tempting or conservative datasets will violate safety constraints or result in sub-optimality, both of which are undesirable.
% The SDM problem is a unique feature of offline safe RL problems, meaning that when training on tempting or conservative datasets, staying close to the behavior policy will either violate the safety constraints or lead to sub-optimality, which is neither desirable. 
Therefore, addressing the SDM problem is essential for the development of regularized offline safe RL algorithms.
% More motivation examples are presented in the Appendix.

\subsection{Mitigating the Safe Dataset Mismatch}
\begin{wrapfigure}{R}{0.45\textwidth}
\vspace{-12mm}
\centering
\includegraphics[width=.98\linewidth]{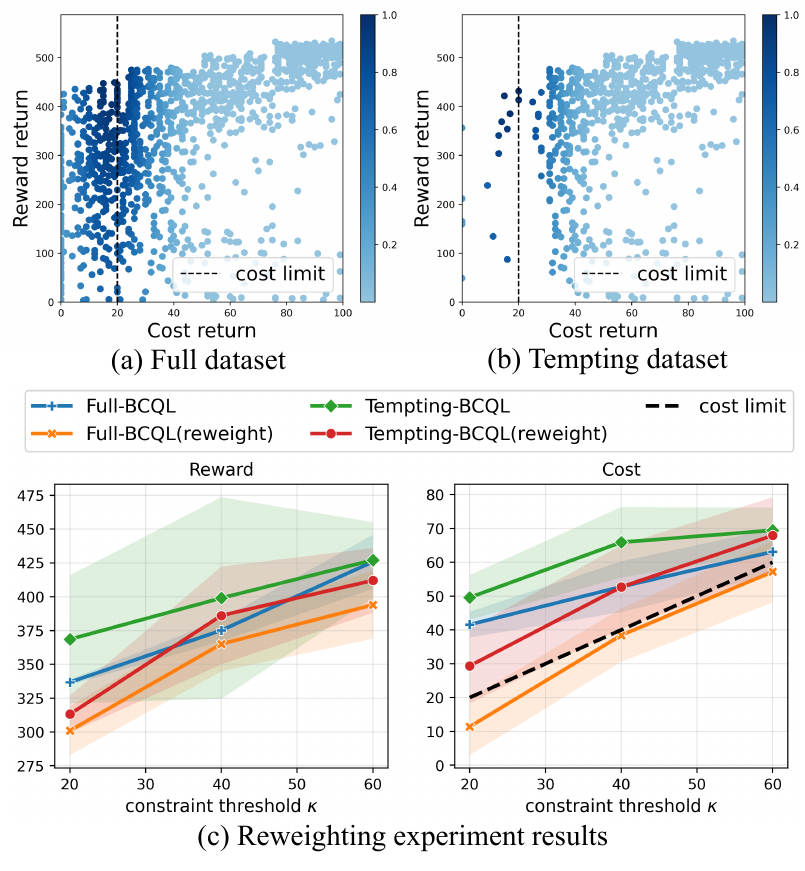}
\vspace{-3mm}
\caption{(a) Reweighting in the dataset with comprehensive coverage. (b) Reweighting in the tempting dataset. (c) Performance evaluation with different weights and datasets.
}
\label{fig: reweighing example}
\vspace{-5mm}
\end{wrapfigure}
\label{subsection: Mitigating the Safe Dataset Mismatch}
We propose to use the distribution shaping of the dataset to mitigate the SDM problem, that is, generating a new dataset $\Dcal_g$ by reshaping the original data distribution. As shown in Figure~\ref{Fig: distribution shaping}, the key idea is to adjust the dataset distribution towards the optimal distribution under $\pi^*$, reducing the distribution discrepancy and mitigating the SDM problem, thus solving the action OOD issue and efficiency-safety balancing problem simultaneously. 

% To mitigate the SDM problem, we propose using the dataset distribution shaping, that is, building a distribution shaping model $M: \Scal \times \Acal \xrightarrow{} \Scal \times \Acal$ such that:
% \begin{equation}
% \label{equ: distribution target}
%     \Dcal_g = M(\Dcal), \quad s.t. \ \tau \sim d_g, \text{for } \tau \in \Dcal_g, \quad D_{TV}(d_g||d^*) \leq \eta, 
%     \quad \text{\textcolor{red}{TODO: modify it}}
% \end{equation}
% where $\Dcal_g$ is a new dataset after distribution shaping, and $d_g(\tau)$ is its corresponding data distribution, $D_{TV}(\cdot | \cdot)$ is the total variation distance between two distributions, and $\eta$ is a tunable threshold for shaping. 
% As shown in Figure~\ref{fig: Conditional Diffusion Model}, the idea of distribution shaping is to adjust the dataset distribution towards the optimal distribution under $\pi^*$, unifying the objectives to solve the challenging issues mentioned in Section~\ref{section: problem formulation}, thus mitigating the SDM problem. 

Dataset reweighing, which assigns different sampling weights to different data points, is a straightforward way to do distribution shaping~\cite{hong2023beyond, hong2023harnessing}. In the offline RL domain, researchers proposed methods to assign high weights to data that achieve high rewards and superior performance in many RL tasks~\cite{hong2023beyond, hong2023harnessing}.
% In contrast to the unconstrained offline RL tasks that only consider reward, Offline safe RL reweighing needs to take both reward and cost information into consideration. 
To validate this idea, we deploy a Boltzmann energy function considering both the reward and the cost for the reweighing strategy to solve the problem (see Appendix~\ref{appendix-sec: sup_exp} for details). The experimental results, shown in Figure~\ref{fig: reweighing example}, validate the effectiveness of this distribution shaping method when the coverage of the dataset is complete.

However, for a more general case where we can only access the low-quality dataset (e.g., tempting datasets in Figure~\ref{fig: reweighing example}), simply performing data reweighing does not work well due to the absence of necessary data. Thus, we propose to use a conditional generative model for more flexible distribution shaping, which generates new data by stitching sub-optimal trajectories for offline training.

% To solve the offline safe RL problem, 
\begin{figure}[t]
    \centering
    % \vspace{-8pt} 
    \includegraphics[width=0.96\linewidth]{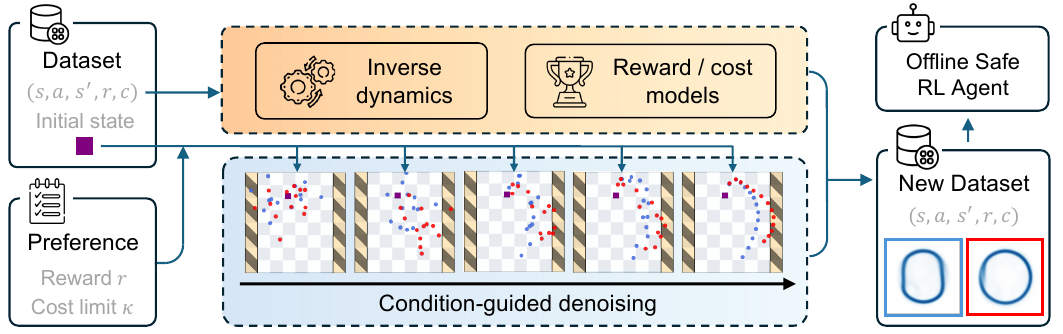}
    \vspace{-1mm}
    \caption{OASIS overview.}
    \label{fig: Conditional Diffusion Model}
    \vspace{-5mm}
\end{figure}

\subsection{Constraint-Conditioned Diffusion Model as Data Generator}

To overcome the limitation of reweighing methods, we propose using diffusion models to generate the dataset that fits the target cost limit to achieve distribution shaping. In the following, we introduce the details of the training and generation phases.

\textbf{Training.} In previous works~\cite{liang2023adaptdiffuser, janner2022planning}, the trajectory planning in offline RL can be viewed as the sequential data generation: 
% \TODO{notation on sub-trajectory, $\tau$ is used in sec. 3.1}:
$
% \boldsymbol{\tau}=\left[\begin{array}{llll}
% s_0 & s_1 & \ldots & s_{L-1} 
% \end{array}\right],
\boldsymbol{\tau}=[s_0, s_1, \ldots, s_{L-1}],
$
where $\boldsymbol{\tau}$ is a subsequence of trajectory with length $L$. Denote $\boldsymbol{x}_k(\boldsymbol{\tau})$ and $\boldsymbol{y}(\boldsymbol{\tau})$ as the $k$-step denoising output of the diffusion model and the denoising conditions such as reward and cost returns, respectively. Then the forward diffusion process is to add noise to $\boldsymbol{x}_k(\boldsymbol{\tau})$ and gradually convert it into Gaussian noise:
\begin{equation}
q\left(\boldsymbol{x}_k(\tau) \mid \boldsymbol{x}_{k-1}(\tau)\right):=\mathcal{N}\left(\boldsymbol{x}_k(\tau) ; \sqrt{1-\beta_k} \boldsymbol{x}_{k-1}(\tau), \beta_k \boldsymbol{I}\right), \ k = 1, ..., K
\end{equation}
where $\beta_k$ is a pre-defined beta schedule, $K$ is the total denoising timestep. Then the trainable denoising step aims at gradually converting the Gaussian noise back to a valid trajectory:
\begin{equation}
p_\theta\left(\boldsymbol{x}_{k-1}(\tau) \mid \boldsymbol{x}_{k}(\tau), \boldsymbol{y}(\boldsymbol{\tau}) \right):=\mathcal{N}\left(\boldsymbol{x}_{k-1}(\tau) \mid \mu_\theta\left(\boldsymbol{x}_k(\tau), \boldsymbol{y}(\tau), k\right), \Sigma_k\right),
\label{equ: inverse sampling}
\end{equation}
where $\theta$ is the trainable parameter. We use a simplified surrogate loss~\cite{ho2020denoising} for optimization:
\begin{equation}
\mathcal{L}_{\text {denoise }}:=\mathbb{E}_{\boldsymbol{x}_0(\tau) \sim q, \epsilon \sim \mathcal{N}(\mathbf{0}, \boldsymbol{I})}\left[\left\|\epsilon-\epsilon_\theta\left(\boldsymbol{x}_k(\tau), \boldsymbol{y}(\tau), k\right)\right\|^2\right].
\label{equ: denoising}
\end{equation}
% \subsubsection{Constraint-conditional diffusion model for dataset shaping}
In this work, we use the classifier-free guidance~\cite{ho2022classifier} for conditional data generation. The condition $\boldsymbol{y}(\tau)$ in (\ref{equ: inverse sampling}) and (\ref{equ: denoising}) is set to $\boldsymbol{y}(\tau) = [C(\tau), R(\tau)]$. Thus, the denoising process depends on the target reward and cost returns of the planned subtrajectory. During training time, the diffusion model learns both an unconditional denoising core $\epsilon_\theta\left(\boldsymbol{x}_k(\tau), \varnothing, k\right)$ and a conditional denoising core $\epsilon_\theta\left(\boldsymbol{x}_k(\tau), \boldsymbol{y}(\tau), k\right)$. We adopt masking~\cite{ajay2022conditional} for the training to zero out the condition of one training trajectory and categorize it as the $\varnothing$ class with probability $0<p<1$. Within the given raw dataset, we also train an inverse dynamics model $\hat{f}: \Scal \times \Scal \xrightarrow{} \Acal$, and reward and cost models $\hat{r}(s, a, s^{\prime}), \hat{c}(s, a, s^{\prime}): \Scal \times \Acal \times \Scal \xrightarrow{} \mathbb{R}$ for labeling. 

\textbf{Generation.} After obtaining a trained model, the next step is to generate a new dataset following the conditions. For diffusion model inference, the denoising core $\epsilon_\theta\left(\boldsymbol{x}_k(\tau), \boldsymbol{y}, k\right)$ is calculated by:
\begin{equation}
    \epsilon_\theta \left(\boldsymbol{x}_k(\tau), \boldsymbol{y}(\tau), k\right) = \epsilon_{\theta} \left(\boldsymbol{x}_k(\tau), \varnothing, k\right) + w_\alpha \ (\epsilon_{\theta} \left(\boldsymbol{x}_k(\tau), \boldsymbol{y}_c(\tau), k\right) - \epsilon_{\theta} \left(\boldsymbol{x}_k(\tau), \varnothing, k\right)),
    \label{equ: classifier-free}
\end{equation}
\begin{wrapfigure}{R}{0.48\textwidth}
\begin{minipage}{0.48\textwidth}
\vspace*{-0.1in}
\begin{algorithm}[H]
\caption{\method\ (generation) }
{\bfseries Input:} \raggedright Raw dataset $\Dcal$, constraint threshold $\kappa$ \par
{\bfseries Output:} \raggedright Generated sub-trajectory $\tau_g$ \par
\begin{algorithmic}[1] % The number tells where the line numbering should start
\STATE Sample a initial state: $s \sim \mathcal{D}$
\STATE Get initial noisy sub-trajectory: $x_k = [s, s_1, ..., s_{L-1}], s_i \sim \mathcal{N}(\boldsymbol{0}, \boldsymbol{I})$ 
\STATE Determine the condition $\boldsymbol{y}_c$;
\FOR{$k=K, ..., 1$}
\STATE Calculate $\epsilon \left(\boldsymbol{x}_k, \boldsymbol{y}_c, k\right)$ (\ref{equ: classifier-free});
\STATE Inverse sampling state $\boldsymbol{x}_{k-1}$ (\ref{equ: inverse sampling})
\ENDFOR
\STATE Get actions, rewards, and costs from $\boldsymbol{x}_{0}$;
% \STATE $\triangleright$ \textit{\textcolor{blue}{Gradient shaping ends}}
\STATE {\bfseries Return:} trajectory $\tau_g$
% \STATE \TODO{notation, batch operation?}
\end{algorithmic} \label{algo: method}
\end{algorithm}
\end{minipage}
\vspace*{-0.4in}
\end{wrapfigure}
where $w_\alpha > 0$ is a constant guidance scale and $\boldsymbol{y}_c := [\hat{C}, \hat{R}]$ is the generation condition. In practice, condition $\hat{C}$ is determined by the cost threshold $\kappa$, and condition $\hat{R}$ is selected based on the raw data distribution. It should be noticed that for guided generation, we fix the initial state, which means that we replace the initial state of each $k$-step noised trajectory as $\boldsymbol{x}_{k}[0] = \boldsymbol{x}_{0}[0]$.

After generating one subtrajectory $\tau_g = \boldsymbol{x}_{0}$, we can get the state and action sequence $s_g = \tau_g[:-1], s_g^{\prime} = \tau_g[1:], a_g = \hat{f}(s_g, s_g^{\prime})$, then label the data $r_g = \hat{r}(s_g, a_g, s_g^{\prime}), c_g = \hat{c}(s_g, a_g, s_g^{\prime})$. Finally, we get a generated dataset $\mathcal{D}_g = \{s_g, a_g, s_g^{\prime}, r_g, c_g\}$ with $|\tau_g|-1$ transition pairs. With this new dataset, we can further train offline safe RL agents. 

In this work, we consider \texttt{BCQ-Lag}~\cite{fujimoto2019off, stooke2020responsive} as the base offline safe RL algorithm. The process of generating one subtrajectory $\tau_g$ is summarized in Alg.~\ref{algo: method}. More details of the implementation are available in Appendix~\ref{appendix-sec: implementation}.

\subsection{Theoritical analysis}
\label{subsection: theoretical analysis}

% When the generated dataset satisfies the target conditions in (\ref{equ: distribution target}), 
We first investigate how the distribution mismatch degrades the policy performance on constraint satisfaction. Suppose that the maximum one-step cost is $C_{\text{max}}=\max_{s,a} c(s,a)$. Based on Lemma 6 in \cite{xu2020error} and Lemma 2 in \cite{cen2024learning}, the performance gap between the policy $\pi$ learned with the dataset $\Dcal$ and the optimal policy is bounded by
\begin{equation}
\label{eq:lemma}
    |V_c^{\pi}(\mu_0) - V_c^*(\mu_0)|\leq \frac{2C_{\text{max}}}{1-\gamma}\TV(d^\Dcal(s) \| d^*(s))
    + \frac{2C_{\text{max}}}{1-\gamma}\E_{d^*(s)}[\TV(\pi(a|s) \| \pi^*(a|s))],
\end{equation}
where $d^\Dcal(s),d^*(s)$ denote the stationary state distribution of the dataset and optimal policy. The proof is given in Appendix~\ref{sec:proof_lemma}. Therefore, a significant mismatch between the dataset and the optimal policy results in both a substantial state distribution TV distance and a large policy shift from the optimal one, which can cause notable performance degradation, especially when the offline RL algorithm enforces the learned policy to closely resemble the behavior policy of the offline data.

Then we provide a theoretical analysis of how our method mitigates this mismatch issue by shrinking the distribution gap, which provides a guarantee of the safety performance of the regularized offline safe RL policy. Let $d_g(s| \boldsymbol{y})$ denote the state marginal of the generated data with condition $\boldsymbol{y}$. We first make the following assumptions.

\begin{assumption}[Score estimation error of the state marginal]\label{ass:score}
There exists a condition $\boldsymbol{y}^*$ such that the score function error of the state marginal is bounded by
\begin{equation}
% \small
    \E_{d^*(s)}\| \nabla_s\log d_g(s|\boldsymbol{y}^*) - \nabla_s \log d^*(s)\| \leq \varepsilon_{\text{score}}^2,
\end{equation}
where $d^*(s)$ is the stationary state distribution induced by the optimal policy $\pi^*$.
\end{assumption}
This assumption is also adopted in previous work~\cite{lee2022convergence, chen2022sampling}. For simplicity, we omit the condition $\boldsymbol{y}^*$ in the following analysis and use $d_g(s), d_g(s,a)$ to denote the generated state or the state-action distribution with condition $\boldsymbol{y}^*$. As we use inverse dynamics $f(a|s,s')$ to calculate actions based on the generated state sequence, the quality of the dataset is also determined by the inverse dynamics. Therefore, we further make the following assumption.

\begin{assumption}[Error of inverse policy]\label{ass:inverse}
The error of action distribution generated by the inverse dynamics is bounded by
\begin{equation}
    \E_{d^*(s)} \left[\KL(\hat{\pi}_{\text{inv}}(\cdot|s) \| \pi^*(\cdot|s))\right] \leq \varepsilon_{\text{inv}},
\end{equation} 
where $\hat{\pi}_{\text{inv}}(a|s)=\E_{s'}[\hat{f}(a|s,s')]$ denotes the empirical inverse policy, which is a marginal of inverse dynamics over $s'$.
\end{assumption}
% In practice, the error $\varepsilon_{\text{inv}}$ depends on the data for inverse policy learning.
Then the distance of generated data distribution to the optimal one is bounded as:
\begin{theorem}[Distribution shaping error bound]
\label{theorem: distribution shaping error}
Suppose that the optimal stationary state distribution satisfies that 1) its score function $\nabla_{s} \log d^*(s)$ is $L$-Lipschitz and 2) its second momentum is bounded. Under Assumption~\ref{ass:score} and \ref{ass:inverse}, the gap of generated state-action distribution to the optimal stationary state-action distribution is bounded by
\begin{equation}
    \TV\left(d_g(s,a)\|d^*(s,a)\right) \leq \tilde{\Ocal}\left(\varepsilon_{\text{score}}\sqrt{K}\right) + \sqrt{\varepsilon_{\text{inv}} / 2} + C(d^*(s), L, K),
\end{equation}
where $C(d^*(s), L, K)$ represents a constant determined by $d^*(s), L$ and $K$.
\end{theorem}
The proof is given in Appendix~\ref{sec:proof_dist_error}. Theorem~\ref{theorem: distribution shaping error} indicates that using the proposed \method\ method, we can shape the dataset distribution towards a bounded neighborhood of the optimal distribution.

Given the generated data, we will then train a regularized offline safe RL policy by (\ref{equ: offline safe RL}). Notice that the regularization term in the objective function in (\ref{equ: offline safe RL}) is equivalent to an explicit policy constraint, and the coefficient $w$ is the corresponding dual variable. Therefore, we make the following assumption on the distance between the learned policy $\pi_\phi$ and the behavior policy.

\begin{assumption}
\label{ass:reg}
Denote the generated dataset as $\Dcal_g$ and the corresponding behavior policy as $\pi_g$, given a fixed coefficient $w$, for the policy $\pi_\phi$ optimized by (\ref{equ: offline safe RL}), there exists a $\varepsilon_{\text{reg}}$ such that
\begin{equation}
\E_{d_g(s)}\left[\KL\left(\pi_\phi(\cdot|s)\|\pi_g(\cdot|s)\right)\right] \leq \varepsilon_{\text{reg}}.
\end{equation}
\end{assumption}
Based on the above assumptions, we can derive the bound of constraint violation of the policy learned on the offline data generated by \method. The proof is given in Appendix~\ref{sec:proof_violation_bound}. 

\begin{theorem}[Constraint violation bound]
\label{thm:violation_bound} For policy $\pi_\phi$ optimized by regularized-based offline safe RL on generated dataset $\Dcal_g$, under assumption~\ref{ass:score} , \ref{ass:inverse} and \ref{ass:reg}, the constraint violation of the trained policy is bounded as:
\begin{equation}
    V_c^{\pi_\phi}(\mu_0) - \kappa \leq 
    \frac{2C_{\text{max}}}{1-\gamma}\left(
    % \underbrace{
        \tilde{\Ocal}\left(\varepsilon_{\text{score}}\sqrt{K}\right) + C(d^*(s), L, K)
    % }_\text{diffusion model error}
    + 
    % \underbrace{
        \sqrt{\varepsilon_{\text{inv}} / 2}
    % }_\text{inverse model error}
    + 
    % \underbrace{
        \sqrt{\varepsilon_{\text{reg}} / 2}
    % }_\text{offline policy constraint}
    \right)
    ,
\end{equation}
where $C(d^*(s), L, K)$ represents a constant determined by $d^*(s), L$ and $K$.
\end{theorem}

\vspace{-5pt}
\section{Experiment}
\vspace{-5pt}
\begin{table}[bp]
  \centering
  \vspace{-6mm}
  \caption{Evaluation results of the normalized reward and cost. The cost threshold is 1. $\uparrow$: the higher the reward, the better. $\downarrow$: the lower the cost (up to threshold 1), the better. \textbf{Bold}: Safe agents whose normalized cost is smaller than 1. {\color[HTML]{656565} Gray}: Unsafe agents. {\color[HTML]{0000FF} \textbf{Blue}}: Safe agent with the highest reward. }
  \resizebox{1.\linewidth}{!}{
  \begin{tabular}{cccccccc}
    \toprule
    \multirow{2}[4]{*}{Algorithm} & \multirow{2}[4]{*}{Stats} & \multicolumn{6}{c}{Tasks} \\
\cmidrule{3-8}          &       & BallRun & CarRun & DroneRun & BallCircle & CarCircle & DroneCircle \\
    \midrule
    \multirow{2}[1]{*}{BC} & reward $\uparrow$ & \textcolor[rgb]{ .455,  .455,  .455}{0.55 ± 0.23} & \textcolor[rgb]{ .455,  .455,  .455}{0.94 ± 0.02} & \textcolor[rgb]{ .455,  .455,  .455}{0.62 ± 0.11 } & \textcolor[rgb]{ .455,  .455,  .455}{0.73 ± 0.05 } & \textcolor[rgb]{ .455,  .455,  .455}{0.59 ± 0.11 } & \textcolor[rgb]{ .455,  .455,  .455}{0.82 ± 0.01} \\
          & cost $\downarrow$  & \textcolor[rgb]{ .455,  .455,  .455}{2.04 ± 1.32} & \textcolor[rgb]{ .455,  .455,  .455}{1.50 ± 1.11} & \textcolor[rgb]{ .455,  .455,  .455}{3.48 ± 0.68} & \textcolor[rgb]{ .455,  .455,  .455}{2.53 ± 0.15} & \textcolor[rgb]{ .455,  .455,  .455}{3.39 ± 0.85} & \textcolor[rgb]{ .455,  .455,  .455}{3.29 ± 0.18} \\
    \multirow{2}[0]{*}{CPQ} & reward $\uparrow$ & \textcolor[rgb]{ .455,  .455,  .455}{0.25 ± 0.11} & \textcolor[rgb]{ .455,  .455,  .455}{0.63 ± 0.51} & \textcolor[rgb]{ .455,  .455,  .455}{0.13 ± 0.30 } & \textbf{0.39 ± 0.34 } & \textbf{0.64 ± 0.02 } & \textcolor[rgb]{ .455,  .455,  .455}{0.01 ± 0.02} \\
          & cost $\downarrow$  & \textcolor[rgb]{ .455,  .455,  .455}{1.34 ± 1.32} & \textcolor[rgb]{ .455,  .455,  .455}{1.43 ± 1.82} & \textcolor[rgb]{ .455,  .455,  .455}{2.29 ± 1.98} & \textbf{0.73 ± 0.66} & \textbf{0.12 ± 0.19} & \textcolor[rgb]{ .455,  .455,  .455}{3.16 ± 3.85} \\
    \multirow{2}[0]{*}{COptiDICE} & reward $\uparrow$ & \textcolor[rgb]{ .455,  .455,  .455}{0.63 ± 0.04} & \textbf{0.90 ± 0.03} & \textcolor[rgb]{ .455,  .455,  .455}{0.71 ± 0.01 } & \textcolor[rgb]{ .455,  .455,  .455}{0.73 ± 0.02 } & \textcolor[rgb]{ .455,  .455,  .455}{0.52 ± 0.01} & \textbf{0.35 ± 0.02 } \\
          & cost $\downarrow$  & \textcolor[rgb]{ .455,  .455,  .455}{3.13 ± 0.17} & \textbf{0.28 ± 0.24} & \textcolor[rgb]{ .455,  .455,  .455}{3.87 ± 0.08} & \textcolor[rgb]{ .455,  .455,  .455}{2.83 ± 0.23} & \textcolor[rgb]{ .455,  .455,  .455}{3.56 ± 0.16} & \textbf{0.12 ± 0.10} \\
    \multirow{2}[0]{*}{BEAR-Lag} & reward $\uparrow$ & \textcolor[rgb]{ .455,  .455,  .455}{0.65± 0.08} & \textcolor[rgb]{ .455,  .455,  .455}{0.55 ± 0.62} & \textcolor[rgb]{ .455,  .455,  .455}{0.10 ± 0.33} & \textcolor[rgb]{ .455,  .455,  .455}{0.89 ± 0.02 } & \textcolor[rgb]{ .455,  .455,  .455}{0.80 ± 0.08 } & \textcolor[rgb]{ .455,  .455,  .455}{0.89 ± 0.04 } \\
          & cost $\downarrow$  & \textcolor[rgb]{ .455,  .455,  .455}{4.38 ± 0.28} & \textcolor[rgb]{ .455,  .455,  .455}{8.44 ± 0.62} & \textcolor[rgb]{ .455,  .455,  .455}{3.72 ± 3.22} & \textcolor[rgb]{ .455,  .455,  .455}{2.84 ± 0.28} & \textcolor[rgb]{ .455,  .455,  .455}{2.89 ± 0.84} & \textcolor[rgb]{ .455,  .455,  .455}{4.03 ± 0.51} \\
    \multirow{2}[0]{*}{BCQ-Lag} & reward $\uparrow$ & \textcolor[rgb]{ .455,  .455,  .455}{0.51 ± 0.19} & \textcolor[rgb]{ .455,  .455,  .455}{0.96 ± 0.06} & \textcolor[rgb]{ .455,  .455,  .455}{0.76 ± 0.07} & \textcolor[rgb]{ .455,  .455,  .455}{0.76 ± 0.04 } & \textcolor[rgb]{ .455,  .455,  .455}{0.79 ± 0.02 } & \textcolor[rgb]{ .455,  .455,  .455}{0.88 ± 0.04 } \\
          & cost $\downarrow$  & \textcolor[rgb]{ .455,  .455,  .455}{1.96 ± 0.88} & \textcolor[rgb]{ .455,  .455,  .455}{2.31 ± 3.22} & \textcolor[rgb]{ .455,  .455,  .455}{5.19 ± 1.08} & \textcolor[rgb]{ .455,  .455,  .455}{2.62 ± 0.29} & \textcolor[rgb]{ .455,  .455,  .455}{3.25 ± 0.28} & \textcolor[rgb]{ .455,  .455,  .455}{3.90 ± 0.55} \\
    \multirow{2}[0]{*}{CDT} & reward $\uparrow$ & \textcolor[rgb]{ .455,  .455,  .455}{0.35 ± 0.01} & \textcolor[rgb]{ 0,  .165,  .835}{\textbf{0.96 ± 0.01}} & \textcolor[rgb]{ .455,  .455,  .455}{0.84 ± 0.12} & \textcolor[rgb]{ .455,  .455,  .455}{0.73 ± 0.01 } & \textcolor[rgb]{ .455,  .455,  .455}{0.71 ± 0.01} & \textcolor[rgb]{ .455,  .455,  .455}{0.17 ± 0.08 } \\
          & cost $\downarrow$  & \textcolor[rgb]{ .455,  .455,  .455}{1.56 ± 1.10} & \textcolor[rgb]{ 0,  .165,  .835}{\textbf{0.67 ± 0.03}} & \textcolor[rgb]{ .455,  .455,  .455}{7.56 ± 0.33} & \textcolor[rgb]{ .455,  .455,  .455}{1.36 ± 0.03} & \textcolor[rgb]{ .455,  .455,  .455}{2.39 ± 0.15} & \textcolor[rgb]{ .455,  .455,  .455}{1.08 ± 0.62} \\
    \multirow{2}[0]{*}{FISOR} & reward $\uparrow$ & \textbf{0.17 ± 0.03} & \textbf{0.85 ± 0.02} & \textcolor[rgb]{ .455,  .455,  .455}{0.44 ± 0.14} & \textbf{0.28 ± 0.03 } & \textbf{0.24 ± 0.05 } & \textbf{0.49 ± 0.05 } \\
          & cost $\downarrow$  & \textbf{0.04 ± 0.06} & \textbf{0.15 ± 0.20} & \textcolor[rgb]{ .455,  .455,  .455}{2.52 ± 0.61} & \textbf{0.00 ± 0.00} & \textbf{0.15 ± 0.27} & \textbf{0.02 ± 0.03} \\
    \multirow{2}[0]{*}{CVAE-BCQL} & reward $\uparrow$ & \textcolor[rgb]{ .455,  .455,  .455}{0.25 ± 0.02} & \textbf{0.88 ± 0.00} & \textcolor[rgb]{ .455,  .455,  .455}{0.21 ± 52.07} & \textcolor[rgb]{ .455,  .455,  .455}{0.49 ± 0.03} & \textcolor[rgb]{ .455,  .455,  .455}{0.60 ± 0.05} & \textcolor[rgb]{ .455,  .455,  .455}{0.01 ± 0.02} \\
          & cost $\downarrow$  & \textcolor[rgb]{ .455,  .455,  .455}{1.40 ± 0.35} & \textbf{0.00 ± 0.00} & \textcolor[rgb]{ .455,  .455,  .455}{2.80 ± 0.63} & \textcolor[rgb]{ .455,  .455,  .455}{1.39 ± 0.27} & \textcolor[rgb]{ .455,  .455,  .455}{1.77 ± 0.47} & \textcolor[rgb]{ .455,  .455,  .455}{3.31 ± 1.66} \\
    \multirow{2}[1]{*}{\method\ (ours)} & reward $\uparrow$ & \textcolor[rgb]{ 0,  .165,  .835}{\textbf{0.28 ± 0.01}} & \textbf{0.85 ± 0.04} & \textcolor[rgb]{ 0,  .165,  .835}{\textbf{0.13 ± 0.08}} & \textcolor[rgb]{ 0,  .165,  .835}{\textbf{0.70 ± 0.01 }} & \textcolor[rgb]{ 0,  .165,  .835}{\textbf{0.76 ± 0.03 }} & \textcolor[rgb]{ 0,  .165,  .835}{\textbf{0.60 ± 0.01 }} \\
          & cost $\downarrow$  & \textcolor[rgb]{ 0,  .165,  .835}{\textbf{0.79 ± 0.37}} & \textbf{0.02 ± 0.03} & \textcolor[rgb]{ 0,  .165,  .835}{\textbf{0.79 ± 0.54}} & \textcolor[rgb]{ 0,  .165,  .835}{\textbf{0.45 ± 0.14}} & \textcolor[rgb]{ 0,  .165,  .835}{\textbf{0.89 ± 0.59}} & \textcolor[rgb]{ 0,  .165,  .835}{\textbf{0.25 ± 0.10}} \\
    \bottomrule
    \end{tabular}%
    \vspace{-5mm}
  }
  \label{tab:addlabel}%
\end{table}%
\label{section: experiment}
In the experiments, we answer these questions: (1) How does the distribution of the dataset influence the performance of regularized offline safe RL? (2) How does our proposed distribution shaping method perform in offline safe RL tasks? (3) How well does the conditional data generator shape the dataset distribution? To address these questions, we set up the following experiment tasks.

\textbf{Environments.} We adopt the continuous robot locomotion control tasks in the public benchmark \texttt{Bullet-Safety-Gym}~\cite{gronauer2022bullet} for evaluation, which is commonly used in previous works~\cite{guo2024temporal, liu2023constrained,zhang2020first}. We consider two tasks, \texttt{Run} and \texttt{Circle}, and three types of robots, \texttt{Ball}, \texttt{Car}, and \texttt{Drone}. We name the environment as \texttt{Agent}-\texttt{Task}. 
% The reward signal is continuous, and the cost signal is binary (sparse).
A detailed description is available in the Appendix~\ref{appendix-sec: sup_exp}.

\textbf{Datasets.} Our experiment tasks are mainly built upon the offline safe RL dataset \texttt{OSRL}~\cite{liu2023datasets}. To better evaluate the tested algorithms with the challenging SDM problem, we create four different training dataset types, \texttt{full}, \texttt{tempting}, \texttt{conservative}, and \texttt{hybrid}. The \texttt{tempting} dataset contains sparse safe demonstrations, the \texttt{conservative} dataset lacks rewarding data points, and the \texttt{hybrid} dataset has scarcity in the medium-reward, medium-cost trajectories. We set different cost thresholds for different datasets. A detailed description and visualization of the datasets are available in Appendix~\ref{appendix-sec: sup_exp}.

\textbf{Baselines.} We compared our method with five types of baseline methods: (1) Q-learning-based algorithms: \texttt{BCQ-Lag}~\cite{fujimoto2019off, stooke2020responsive}, \texttt{BEAR-Lag}~\cite{kumar2019stabilizing, stooke2020responsive}, and \texttt{CPQ}~\cite{xu2022constraints}; (2) Imitation learning:  Behavior Cloning (\texttt{BC})~\cite{xu2022constraints}; (3) Distribution correction estimation: \texttt{COptiDICE}~\cite{lee2021optidice}, and (4) sequential modeling algorithms: \texttt{CDT}~\cite{liu2023constrained} and \texttt{FISOR}~\cite{zheng2024safe}; (5) Data generate: \texttt{CVAE-BCQL}: we train BCQ-Lag agents on the datasets generated by Conditional Variational Autoencoder (CVAE)~\cite{kingma2014semi}.

\textbf{Metrics.} We use the normalized cost return and the normalized reward return as the evaluation metric for comparison in Table \ref{tab:addlabel} and \ref{tab:ablation}. The normalized cost is defined as $C_{\text{normalized}} = C_\pi/ \kappa$, where $C_\pi$ is the cost return and $\kappa$ is the cost threshold. The agent is safe if $C_{\text{normalized}} \leq 1$. The normalized reward is computed by $R_{\text{normalized}} = R_\pi / r_{\text{max}}(\Mcal)$, where $r_{\text{max}}(\Mcal)$ is the maximum empirical reward return for task $\Mcal$ within the given dataset. We report the averaged results and standard deviations over $3$ seeds for all the quantity evaluations.
% \begin{equation}
%     C_{\text{normalized}} = C_\pi/ \kappa ,
% \end{equation}

\begin{figure}[t]
\vspace{-5mm}
     \centering
     \begin{subfigure}[b]{0.49\textwidth}
         \centering
         \includegraphics[width=\textwidth]{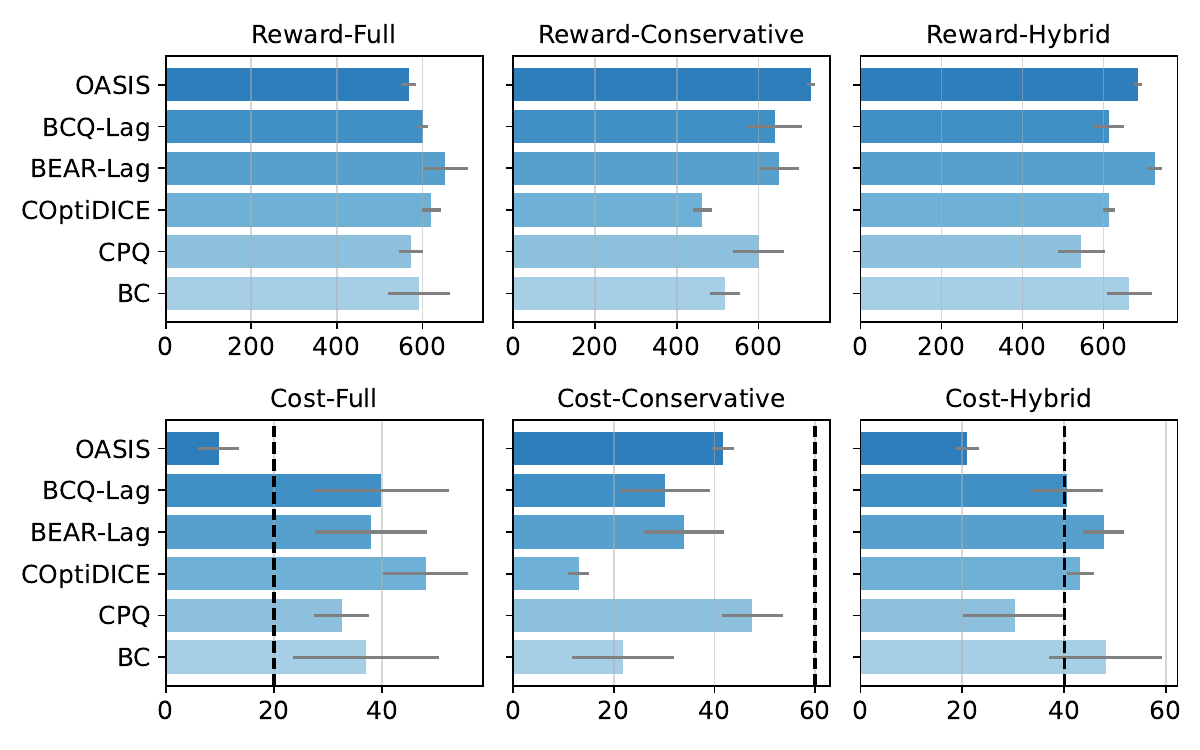}
         \vspace{-6mm}
         \caption{Ball-Circle}
         \label{fig: ballcircle-diff_datasets}
     \end{subfigure}
     \hfill
     \begin{subfigure}[b]{0.49\textwidth}
         \centering
         \includegraphics[width=\textwidth]{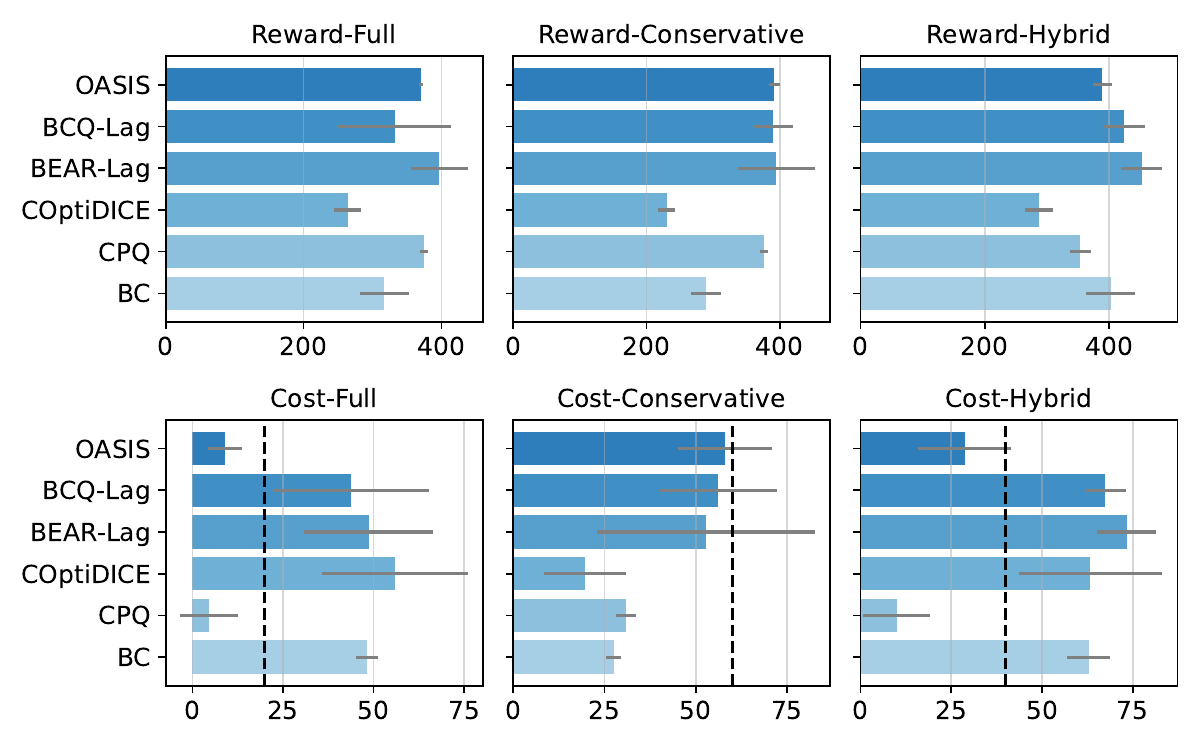}
         \vspace{-6mm}
         \caption{Car-Circle}
         \label{fig: Carcircle-diff_datasets}
     \end{subfigure}
     \vspace{-1mm}
        \caption{Performance with different datasets and varying constraint thresholds.}
        \label{fig: varying dataset}
    \vspace{-6mm}
\end{figure}
\vspace{-3mm}

\subsection{How can conditional data generation benefit offline safe RL?}

\textbf{Performance degradation with SDM problems.} The comparison results on the \texttt{tempting} dataset are presented in Table.~\ref{tab:addlabel} with the cost threshold $\kappa = 20$ before normalization. Results of \texttt{BC} show the mismatch between the behavior policy and the safe policy, as the cost returns significantly violate the safety constraints. The results of \texttt{BCQ-Lag} and \texttt{BEAR-Lag} show this mismatch further influences the regularized-based algorithms, leading to constraint violations. This is because the regularization term pushes the policy towards the unsafe behavior policy. 
\begin{wrapfigure}{R}{0.5\textwidth}
\vspace{-7mm}
\centering
\includegraphics[width=.98\linewidth]{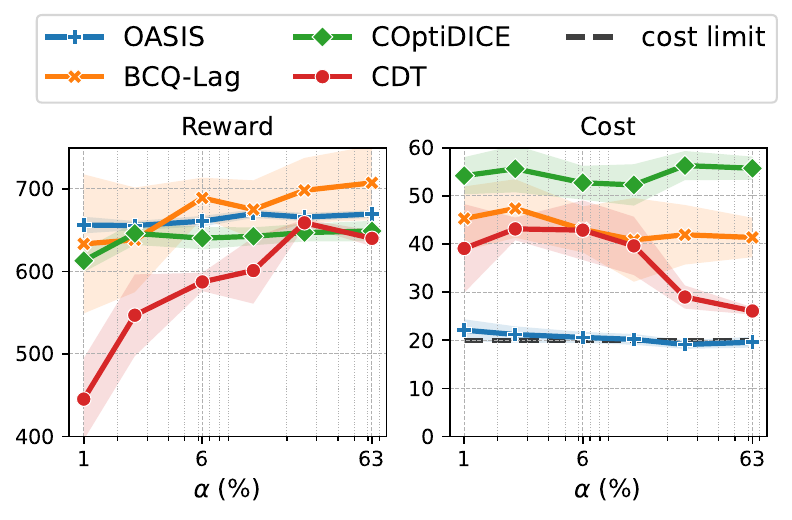}
\vspace{-3mm}
\caption{Data efficiency on the \texttt{Ball-Circle} task with a \texttt{tempting} dataset.
}
\label{fig: exp dataefficiency}
\vspace{-4mm}
\end{wrapfigure}
The conservative Q function estimation method \texttt{CPQ}, exhibits a significant reward degradation in all the tested tasks, which arises from the drawback of the pessimistic estimation methods learning over-conservative behaviors. \texttt{COptiDICE} is also not able to learn safe and rewarding policies, showing that even using distribution correction estimation is not enough to solve the SDM problem. For the sequential modeling algorithms, \texttt{CDT} shows poor safety performance and \texttt{FISOR} tends to be over-conservative with poor reward performance. This is because both methods require a large amount of high-quality data while the trajectories with low cost and high reward are sparse in this task. These observations further motivate the distribution shaping for offline safe RL.
% These results show that imperfection demonstration also has negative impacts on general offline safe RL algorithms.

\textbf{Performance improvement using \method.} From Table~\ref{tab:addlabel}, we find that only our method \method\ can learn safe and rewarding policies by mitigating the SDM problem.
In addition to the results on the \texttt{tempting} dataset, we also provide evaluation results within different types of datasets and constraint thresholds in Figure~\ref{fig: ballcircle-diff_datasets} and Figure~\ref{fig: Carcircle-diff_datasets}. We can observe that most baselines still fail to learn a safe policy within different task conditions due to the SDM issue.
In contrast, our proposed \method\ method achieves the highest reward among all safe agents, which shows strength in more general cases. 
% When trained on \texttt{conservative} datasets, our method generates rewarding data points, reducing the SDM issue and thus improving the final performance of RL agents.  For the \texttt{hybrid} dataset, Our \method\ method successfully distills the information and then generates data with the target domain, thus benefiting the RL agent training.

\textbf{High data efficiency of \method.} We evaluate data efficiency by changing the amount of generated data. Denote $\alpha$ as the amount ratio of the generated data and the raw dataset, the evaluation results are shown in Figure~\ref{fig: exp dataefficiency}, which indicates that we can still learn a good policy using a small amount of high-quality data ($\alpha < 2\%$) generated by \method. In contrast, baseline methods show significant performance degradation when the data are sparse as the noisy data is of low quality.
\begin{figure}[t]
    \centering
    \vspace{-6mm}
    \includegraphics[width=0.98\linewidth]{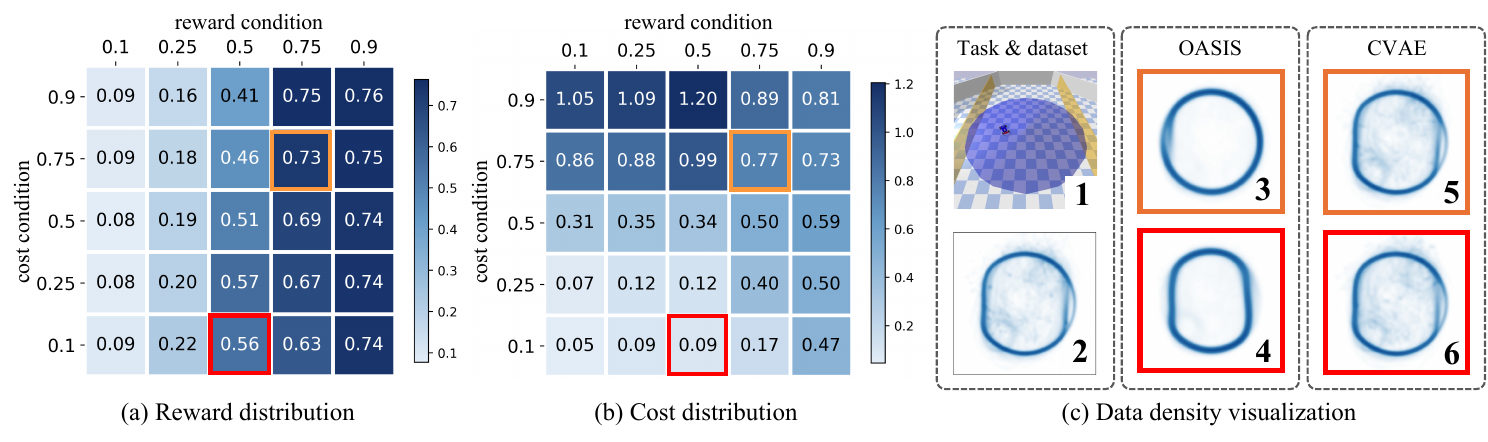}
    \vspace{-1mm}
    \caption{ (a)(b) Reward and cost performance of the generated data: $\mathbb{E} \left[r(s, a)\right], \mathbb{E} \left[c(s, a)\right], (s, a) \sim d_g$. The x-axis and y-axis mean the reward and cost conditions and the values of both conditions and expectations are normalized with the same scale. (c) Visualization of the data density. $1$: \texttt{Car-circle} task; $2$: The density of the $(x, y)$ position of the raw dataset; $3, 4$: The density of the position $(x, y)$ of the data generated by \method\ under conditions $[0.1, 0.5]$ and $[0.75, 0.75]$; $5, 6$: the density of the position $(x, y)$ of the data generated by \texttt{CVAE} under conditions $[0.1, 0.5]$ and $[0.75, 0.75]$}
    \label{fig: density}
    \vspace{-3mm}
\end{figure}
% \subsection{How does data distribution affect the performance of regularized offline safe RL?}
% 1-2 Figures
\subsection{How can \method\ shape the dataset distribution?}
\textbf{Successful distribution shaping.} To show the capability of dataset distribution shaping of the proposed \method\ and baseline \texttt{CVAE}, we generate the dataset under different conditions and visualize them in Figure~\ref{fig: density}. We can first observe that, when using different conditions, the expectations of reward and cost of the generated dataset change accordingly. This shows the strong capability of our method in distribution shaping. We also visualize the density of the generated data. In the \texttt{Car-circle} task, the robot receives high rewards when moving along the circle boundary and receives costs when it exceeds the boundaries on both sides, as shown in Figure~\ref{fig: density}(c). The original dataset contains trajectories with various safety performances. When using a low-cost condition, the generated data are clustered within the safety boundary to satisfy the constraints. When using a high-reward condition, the generated data points are closer to the circle boundary and receive higher rewards. In contrast, the baseline \texttt{CVAE} cannot successfully incorporate the conditions in data generation, resulting in almost similar datasets with different conditions as shown in Figure ~\ref{fig: density}(c).

\begin{wraptable}{r}{0.5\textwidth}
\vspace{-17pt}
\caption{Ablation study on denoising step $K$}\label{wrap-tab:1}
\vspace{-4pt}
\resizebox{1.\linewidth}{!}{
    \begin{tabular}{ccccc}
    \toprule
    \multicolumn{2}{c}{$K$} & 10    & 20    & 40 \\
    \midrule
    Ball- & reward & 0.71 ± 0.02 & 0.70 ± 0.01 & 0.71± 0.01 \\
    Circle & cost  & 0.72 ± 0.10 & 0.45 ± 0.14 & 0.99 ± 0.13 \\
    Ball- & reward & 0.29 ± 0.04 & 0.28 ± 0.01 & 0.29 ± 0.01 \\
    Run   & cost  & 0.16 ± 0.14 & 0.79 ± 0.37 & 0.00 ± 0.00 \\
    \bottomrule
    \end{tabular}%
        }
        \label{tab:ablation}
    \vspace{-10pt}
\end{wraptable}

\subsection{Robust performance against denoising steps}
We conduct an ablation study on the key hyperparameter of the proposed \method\ method. The experiment related to the denoising steps $K$ is presented in Table~\ref{tab:ablation}. Performance does not change much with different values, which shows the robustness of the proposed \method\ method.

\vspace{-5pt}
\section{Conclusion}
\vspace{-5pt}
\label{section: conclusion}
In this paper, we study the challenging problem in offline safe RL: the safe data mismatch between the imperfect demonstration and the target performance requirements. To address this issue, we proposed the \method\ method, which utilizes a conditional diffusion model to realize the dataset distribution shaping, benefiting offline safe RL training. In addition to the theoretical guarantee of performance improvement, we also conduct extensive experiments to show the superior performance of \method\ in learning a safe and rewarding policy on many challenging offline safe RL tasks. More importantly, our method shows good data efficiency and robustness to hyperparameters, which makes it preferable for applications in many real-world tasks.

There are two limitations of \method: 1) Offline training takes longer: our method involves preprocessing the offline dataset to enhance quality, which requires more time and computing resources; 2) Achieving zero-constraint violations remains challenging with imperfect demonstrations. One potential negative social impact is that misuse of our method may cause harmful consequences and safety issues. Nevertheless, we believe that our proposed method can inspire further research in the safe learning community and help to adapt offline RL algorithms to real-world tasks with safety requirements.

% \newpage

% \clearpage
\newpage
\bibliographystyle{unsrt}
\bibliography{neurips}
\medskip
% \clearpage
% \newpage
% \input{checklist}

% \clearpage
\newpage

\appendix
\addcontentsline{toc}{section}{Appendix} % Add the appendix text to the document TOC
\part{Appendix} % Start the appendix part
\parttoc % Insert the appendix TOC

\section{Proofs}
\label{appendix: proof}
\subsection{Proof of Eq.(\ref{eq:lemma})}
\label{sec:proof_lemma}

By definition of the stationary state-action distribution, 
% \begin{lemma}
% Suppose that the maximum one-step cost is $C_{\text{max}}=\max_{s,a} c(s,a)$. $V_c^\pi(\mu_0)=\E_{s_0\sim\mu_0}[V_c^{\pi}(s_0)]$ denotes the cost performance of the policy $\pi$, then 
% \begin{equation}
%     |V_c^{\pi}(\mu_0) - V_c^*(\mu_0)|\leq \frac{2C_{\text{max}}}{1-\gamma}\TV(d^*(s) \| d^\pi(s))
%     + \frac{2C_{\text{max}}}{1-\gamma}\E_{d^*(s)}[\TV(\pi(a|s) \| \pi^*(a|s))],
% \end{equation}
% where $V_c^*(\mu_0)$ is the cost performance of the optimal policy $\pi^*$.
% \end{lemma}
\begin{align}
    % \begin{split}
    &|V_c^{\pi}(\mu_0) - V_c^*(\mu_0)| \\
    = & \frac{1}{1-\gamma}\left|\E_{(s,a)\sim d^{\pi}}[c(s,a)] - \E_{(s,a)\sim d^{*}}[c(s,a)]\right| \\
    \leq & \frac{C_{\text{max}}}{1-\gamma}\sum_{s,a}|d^{\pi}(s,a) - d^*(s,a)|\\
    = & \frac{2C_{\text{max}}}{1-\gamma}\TV(d^{\pi}(s,a) \| d^*(s,a))\\
    \leq & \frac{2C_{\text{max}}}{1-\gamma}\left(\TV(d^{\pi}(s,a) \| d^*(s) \pi(a|s)) + \TV(d^*(s) \pi(a|s) \| d^*(s,a)) \right)\\
    = & \frac{2C_{\text{max}}}{1-\gamma} \TV(d^{\pi}(s) \| d^*(s)) + \frac{2C_{\text{max}}}{1-\gamma}\E_{d^*(s)}[\TV(\pi(a|s) \| \pi^*(a|s))]
    % \end{split}
\end{align}
The second inequality holds by triangle inequality for total variation distance.

In general, the stationary distribution of learned policy is in between of the empirical distribution of offline data $d^\Dcal$ and optimal $d^*$. Therefore, we can obtain
\begin{equation}
\label{eq:performance_lemma}
    |V_c^{\pi}(\mu_0) - V_c^*(\mu_0)|\leq \frac{2C_{\text{max}}}{1-\gamma}\TV(d^*(s) \| d^\Dcal(s))
    + \frac{2C_{\text{max}}}{1-\gamma}\E_{d^*(s)}[\TV(\pi(a|s) \| \pi^*(a|s))].
\end{equation}

\subsection{Proof of Theorem~\ref{theorem: distribution shaping error}}
\label{sec:proof_dist_error}
\begin{proof}
By triangle inequality, we first decompose the TV distance between state-action distributions into a state distribution distance and a policy distance,
\begin{align}
    &\TV\left(d_g(s,a)\|d^*(s,a)\right) \\
    = &\TV\left(d_g(s)\pi_g(a|s)\|d^*(s)\pi^*(a|s)\right)\\
    \leq &\TV\left(d_g(s)\pi_g(a|s)\|d^*(s)\pi_g(a|s)\right) + \TV\left(d^*(s)\pi_g(a|s) \| d^*(s)\pi^*(a|s) \right) \\
    = & \TV\left( d_g(s) \| d^*(s)\right) + \E_{d^*(s)}[\TV(\pi_g(a|s)\| \pi^*(a|s))]
\end{align}

We then consider two parts separately. 

For the stationary state distribution distance, we suppose the optimal distribution $d^*(s)$ has a $L$-Lipschitz smooth score function and bounded second momentum. Meanwhile, note that the score function in Assumption~\ref{ass:score} is closely related to the denoising model $\epsilon_\theta$~\cite{song2019generative, dhariwal2021diffusion}:
\begin{equation}
    \nabla_s \log d_g(s|\boldsymbol{y}) = -\frac{1}{\sqrt{1-\bar{\alpha}_t}} \epsilon_\theta (s|\boldsymbol{y}),
\end{equation}
where $\epsilon_\theta(s|\boldsymbol{y})$ is the state marginal of the practical denoising model in Eq.(\ref{equ: classifier-free}). Therefore, by theorem 2 in~\cite{chen2022sampling}, under Assumption~\ref{ass:score}, we have 
\begin{equation}
    \TV(d_g(s)\|d^*(s))\lesssim \sqrt{\KL(d^*(s)\|\Ncal(\mathbf{0}, \mathbf{I}^{|\Scal|}))}\exp(-K) + L(\sqrt{|\Scal|}+\mathbf{m}_2)\sqrt{K} +\varepsilon_{\text{score}}\sqrt{K}
\end{equation}
where $K$ is the number of denoising timestep, $|\Scal|$ is the dimension of the state space, and $\mathbf{m}_2$ is the second momentum of $d^*(s)$. Therefore, aggregating the first two terms in RHS, we have
\begin{equation}
\label{eq:ddpm_proof_final}
    \TV\left( d_g(s) \| d^*(s) \right) \leq \tilde{\Ocal}\left(\varepsilon_{\text{score}}\sqrt{K}\right)+ C(d^*(s), L, K),
\end{equation}
where $C(\dots)$ is a constant w.r.t $d^*, L, K$.

Regarding the policy distance. By Pinsker's inequality, 
\begin{equation}
\TV(\pi_g(a|s)\| \pi^*(a|s)) \leq \sqrt{\frac{1}{2} \KL(\pi_g(a|s)\| \pi^*(a|s))}
\end{equation}
Meanwhile, since the action is generated by the inverse policy, i.e., $\pi_g=\hat{\pi}_{\text{inv}}$, by Assumption~\ref{ass:inverse}, we have
\begin{align}
\E_{d^*(s)}\left[\TV(\pi_g(a|s)\| \pi^*(a|s))\right] &\leq \E_{d^*(s)}\left[\sqrt{\frac{1}{2} \KL(\pi_g(a|s)\| \pi^*(a|s))}\right] \\
& \leq \sqrt{\frac{1}{2} \E_{d^*(s)}\left[\KL(\pi_g(a|s)\| \pi^*(a|s))\right]}\\
& = \sqrt{\varepsilon_{\text{inv}} / 2} \label{eq:inv_proof_final}
\end{align}
where the second inequality holds by Jensen's inequality.

Combining the Eq.(\ref{eq:ddpm_proof_final}) and Eq.(\ref{eq:inv_proof_final}), we finish the proof of theorem~\ref{theorem: distribution shaping error}.
\end{proof}

\subsection{Proof of Theorem~\ref{thm:violation_bound}}
\label{sec:proof_violation_bound}

We start from the Eq.(\ref{eq:lemma}). The policy distance can be further decomposed into
\begin{equation}
    \TV(\pi(a|s) \| \pi^*(a|s)) \leq \TV(\pi(a|s) \| \pi^\Dcal(a|s)) + \TV(\pi^\Dcal(a|s) \| \pi^*(a|s)).
\end{equation}
By Assumption~\ref{ass:inverse} and \ref{ass:reg} and Jensen's inequality, we have
\begin{align}
    & \E_{d^*(s)}[\TV(\pi(a|s) \| \pi^*(a|s))] \\
    \leq &  \E_{d^*(s)}[\TV(\pi(a|s) \| \pi^\Dcal(a|s))] + \E_{d^*(s)}[\TV(\pi^\Dcal(a|s) \| \pi^*(a|s))] \\
    \leq &\E_{d^*(s)}\left[\sqrt{\KL(\pi(a|s) \| \pi^\Dcal(a|s))/2}\right] + \E_{d^*(s)}\left[\sqrt{\KL(\pi^\Dcal(a|s) \| \pi^*(a|s))/2}\right] \\
    \leq & \sqrt{\E_{d^*(s)}\left[\KL(\pi(a|s) \| \pi^\Dcal(a|s))\right]/2} + \sqrt{\E_{d^*(s)}\left[\KL(\pi^\Dcal(a|s) \| \pi^*(a|s))\right] /2} \\
    \leq & \sqrt{\varepsilon_{\text{reg}}/2} + \sqrt{\varepsilon_{\text{inv}} /2} \label{eq:policy_optimal_gap_final}
\end{align}

Plug-in Eq.(\ref{eq:ddpm_proof_final}) and (\ref{eq:policy_optimal_gap_final}) into Eq.(\ref{eq:performance_lemma}), we have
\begin{equation}
    |V_c^{\pi}(\mu_0) - V_c^*(\mu_0)|\leq \frac{2C_{\text{max}}}{1-\gamma}\left(
    % \underbrace{
        \tilde{\Ocal}\left(\varepsilon_{\text{score}}\sqrt{K}\right) + C(d^*(s), L, K)
    % }_\text{diffusion model error}
    + 
    % \underbrace{
        \sqrt{\varepsilon_{\text{inv}} / 2}
    % }_\text{inverse model error}
    + 
    % \underbrace{
        \sqrt{\varepsilon_{\text{reg}} / 2}
    % }_\text{offline policy constraint}
    \right).
\end{equation}

Meanwhile, notice that the optimal policy is constraint satisfactory, i.e., 
\begin{equation}
    V_c^*(\mu_0) \leq \kappa.
\end{equation}
Therefore, we have
\begin{equation}
    V_c^{\pi}(\mu_0) - \kappa\leq \frac{2C_{\text{max}}}{1-\gamma}\left(
    % \underbrace{
        \tilde{\Ocal}\left(\varepsilon_{\text{score}}\sqrt{K}\right) + C(d^*(s), L, K)
    % }_\text{diffusion model error}
    + 
    % \underbrace{
        \sqrt{\varepsilon_{\text{inv}} / 2}
    % }_\text{inverse model error}
    + 
    % \underbrace{
        \sqrt{\varepsilon_{\text{reg}} / 2}
    % }_\text{offline policy constraint}
    \right),
\end{equation}
which finishes the proof of Theorem~\ref{thm:violation_bound}.

\section{Supplementary experiments}
\label{appendix-sec: sup_exp}
\subsection{Trajectory reweighting for distribution shaping}
In this section, we provide details about the trajectory reweighting experiment presented in section~\ref{subsection: Mitigating the Safe Dataset Mismatch}. Following previous work~\cite{hong2023beyond, hong2023harnessing} in offline RL, we adopted datapoint reweighting in policy optimization, which can be formulated via importance sampling as:
\begin{equation}
\mathcal{J}^{w}_{\text{off}}(\pi, {\lambda}) \approx \mathbb{E}_{(s, a) \sim \mathcal{D}_w}[\mathcal{J}_{\text{off}}(\pi, {\lambda})]=\mathbb{E}_{(s, a) \sim \mathcal{D}}[w(s, a) \mathcal{J}_{\text{off}}(\pi, {\lambda})],
\end{equation}
where $\mathcal{J}^{w}_{\text{off}}(\pi, {\lambda})$ is the objective function after reweighting. In this experiment, we utilize a Boltzmann energy function as adopted in \cite{hong2023beyond} for offline RL tasks:
\begin{equation}
w(\tau) \propto \exp \left(\alpha_1 R(\tau)^2+\alpha_2(C(\tau)-\kappa)^2\right)
\end{equation}
Here $w(\tau)$ means that all the state-action pairs in one trajectory share the same weight, which is related to the cost and reward returns $C(\tau), R(\tau)$. We adopt $\alpha_1 = \alpha_2 = 1$ in the experiments shown in Figure~\ref{fig: reweighing example}. 
% In addition to the experiment shown in Figure~\ref{fig: reweighing example}, we also provide more options for the loss calculation and weighting strategy. \TODO{}

% \subsection{Supplementary Ablation Study}
% \TODO{}

\subsection{Supplementary CVAE data generation results}
\begin{figure}[h]
    \centering
    \vspace{-1mm}
    \includegraphics[width=0.98\linewidth]{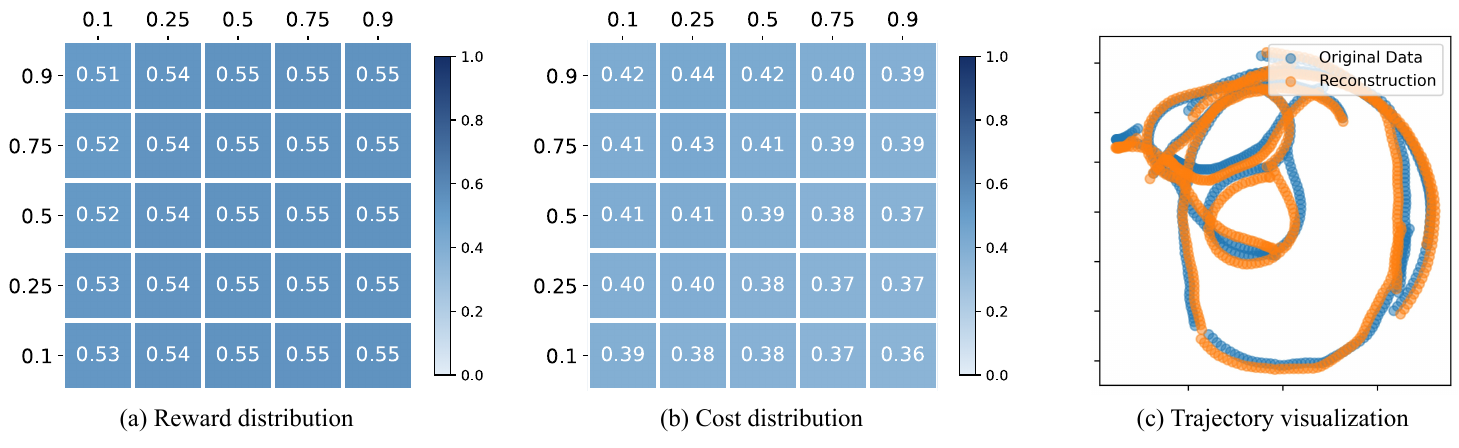}
    \vspace{-1mm}
    \caption{CVAE reconstruction. (a) Reward performance of the generated data: $\mathbb{E} \left[r(s, a)\right], (s, a) \sim d_g$, (b) Cost performance of the generated data: $\mathbb{E} \left[c(s, a)\right], (s, a) \sim d_g$. In (a) and (b), the x-axis and y-axis mean the reward and cost conditions, and the value of both conditions and expectations are normalized to the same scale: $[0, 1]$; (c) The data reconstruction results using the condition $[0.1, 0.5]$ of $10$ sampled trajectories in the dataset.}
    \label{fig: cvae app exp}
\end{figure}

Due to the page limit, we omit the visualization of reward and cost distribution using the CVAE method for data generation. Here we provide the results in Figure~\ref{fig: cvae app exp}. From the reward performance and the cost performance, we can observe that CVAE can hardly encode conditions into the data reconstruction, leading to similar results when setting different conditions. From the trajectory reconstruction results shown in Figure~\ref{fig: cvae app exp}(c), we can observe that the generated trajectories are almost the same as the original one. This feature is not desirable for our distribution shaping purpose. In contrast, our method \method\ can successfully shape the distribution as shown in Figure~\ref{fig: density}, with the strong capability of the diffusion model in the condition-guided denoising process.

\section{Implementation details}
\label{appendix-sec: implementation}

\subsection{Environment details}
% \textbf{Tasks.}
Due to the page limit, we omit some descriptions of experiments in the main context. Here we give the more details about our experiment tasks. Both the Circle task and the Run task are from a publicly available benchmark~\cite{gronauer2022bullet}.

\textbf{Circle tasks.}
The agents are rewarded for running along a circle boundary.
The reward function is defined as:
\begin{equation}
    r(s, a, s^{\prime}) = \frac{-yv_x + xv_y}{1 + |\sqrt{x^2+y^2-radius}|} + r_{robot}(s)
\end{equation}
where $x$, $y$ are the positions of the agent with state $s^{\prime}$, $v_x$, and $v_y$ are velocities of the agent with state $s^{\prime}$. $radius$ is the radius of the circle area, and $r_{robot}(\bm{s_t})$ is the specific reward for different robot.

The agent gets cost when exceeding the boundaries. The cost function is defined as:
\begin{equation}
    \text{Boundary: } c(\bm{s_t}) = \bm{1}(|x| > x_{\text{lim}})
\end{equation}
where $x_{lim}$ is the boundary position.

\textbf{Run tasks.} Agents are rewarded for running fast along one fixed direction and are given costs if they run across the boundaries or exceed a velocity limit.
The reward function is defined as:
\begin{equation}
    r(s, a, s^{\prime})  = ||\bm{x_{t-1}}-\bm{g}||_2 - ||\bm{x}_t-\bm{g}||_2 + r_{robot}(s_t) 
\end{equation}
The cost function is defined as:
\begin{equation}
    c(s, a, s^{\prime})  =  \max(1, \bm{1} (|y| > y_{lim}) + \bm{1}(||\bm{v_t}||_2 > v_{lim}))
\end{equation}
where $v_{lim}$ is the speed limit, and $y_{lim}$ is the $y$ position of the boundary, $\bm{v_t} = [v_x, v_y]$ is the velocity of the agent with state $s^{\prime}$, $\bm{g}= [g_x, g_y]$ is the position of a virtual target, $\bm{x_t} = [x_t, y_t]$ is the position of the agent at timestamp $t$, $\bm{x_{t-1}}$ is the Cartesian coordinates of the agent with state $s$, $\bm{x_{t}}$ is the Cartesian coordinates of the agent with state $s^{\prime}$, and $r_{robot}(\bm{s_t})$ is the specific reward for the robot.

\textbf{Agents.} We use three different robot agents in our experiments: \texttt{Ball}, \texttt{Car}, and \texttt{Drone}. The action space dimension, observation space dimension, and the timesteps for these six tasks are shown in Table.~\ref{tab: env}.

\begin{table}[htbp]
  \centering
  \caption{Environment description}
    \begin{tabular}{cccc}
    \toprule
          & Max   & Action  & Observation \\
          & timestep & space dimension & space dimension \\
    \midrule
    BallRun & 100   & 2     & 7 \\
    CarRun & 200   & 2     & 7 \\
    DroneRun & 200   & 4     & 17 \\
    BallCircle & 200   & 2     & 8 \\
    CarCircle & 300   & 2     & 8 \\
    DroneCircle & 300   & 4     & 18 \\
    \bottomrule
    \end{tabular}%
  \label{tab: env}%
\end{table}%

% \TODO{List $r_{max}, r_{min}$ for normalization?}

\newpage
\subsection{Dataset details}
\begin{figure}[ht]
    \centering
    % \vspace{-8pt} 
    \includegraphics[width=0.98\linewidth]{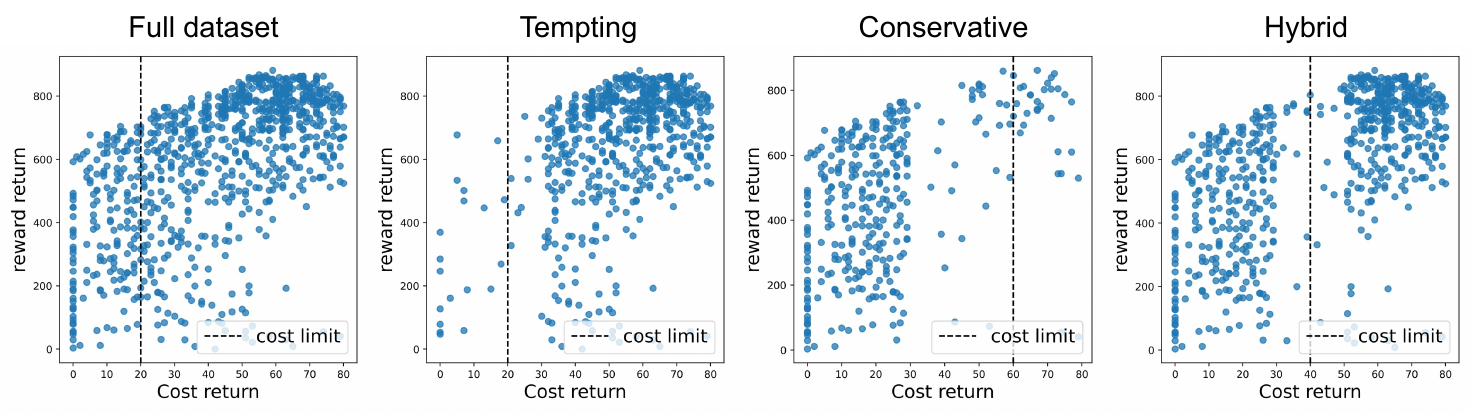}
    \caption{BallCircle Dataset types. Each point represents $(C(\tau), R(\tau))$ of a trajectory $\tau$ in the dataset. }
    \label{Fig: BC exp dataset}
\end{figure}

\begin{figure}[ht]
    \centering
    % \vspace{-8pt} 
    \includegraphics[width=0.98\linewidth]{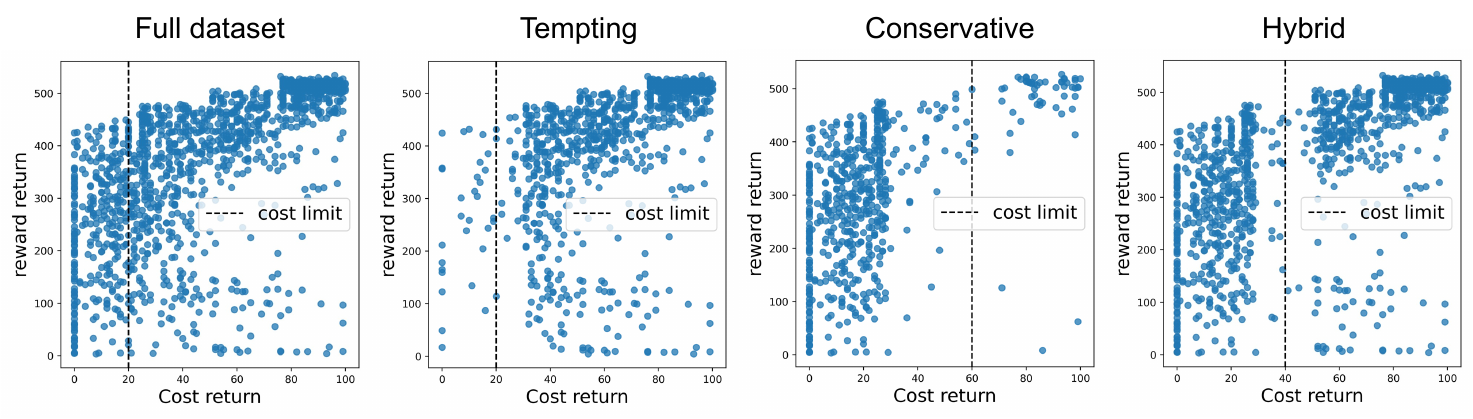}
    \caption{CarCircle Dataset types. Each point represents $(C(\tau), R(\tau))$ of a trajectory $\tau$ in the dataset. }
    \label{Fig: CC exp dataset}
\end{figure}

\begin{figure}[ht]
    \centering
    % \vspace{-8pt} 
    \includegraphics[width=0.73\linewidth]{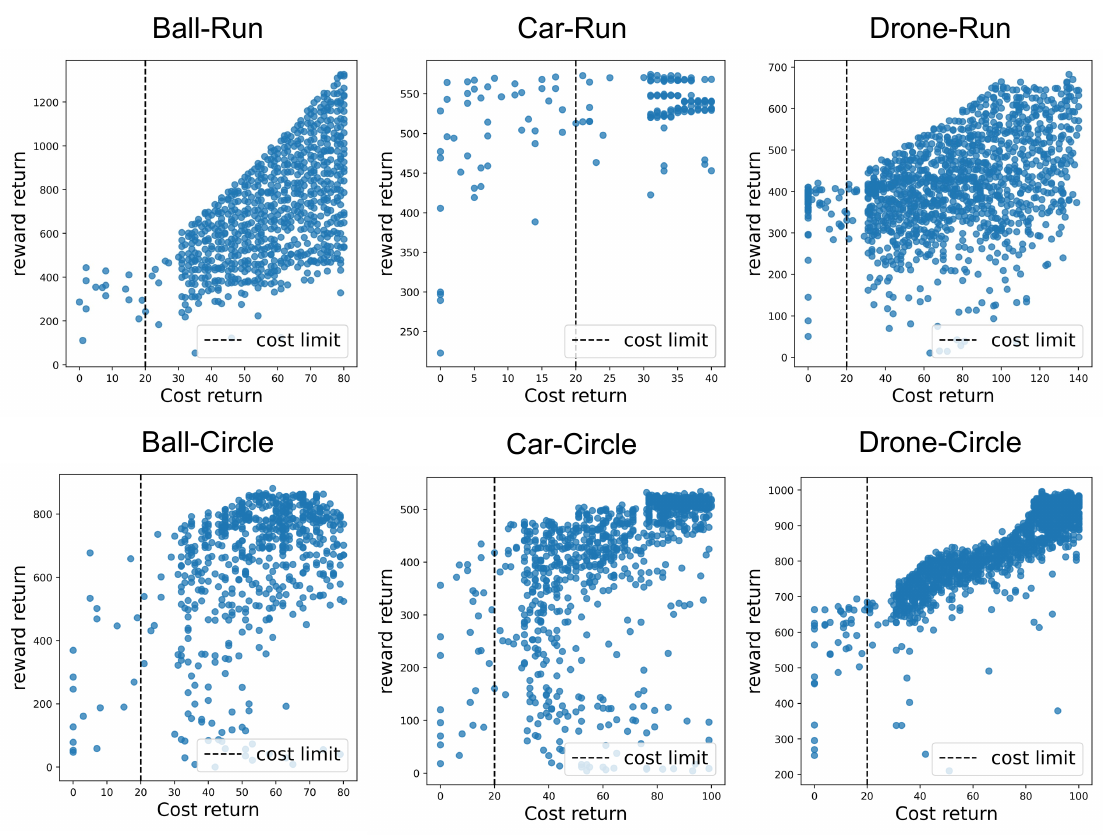}
    \caption{All tempting datasets. Each point represents $(C(\tau), R(\tau))$ of a trajectory $\tau$ in the dataset. }
    \label{Fig: all tempting exp dataset}
\end{figure}

% \textbf{Dataset types.}
We provide details about the dataset types we presented in the experiment part. The \texttt{Full}, \texttt{Tempting}, \texttt{Conservative}, and \texttt{Hybrid} datasets for \texttt{Ball-Circle} and \texttt{Car-Circle} tasks are shown in Figure~\ref{Fig: BC exp dataset}, \ref{Fig: CC exp dataset}, respectively. All the \texttt{Tempting} datasets associated with results in Table~\ref{tab:addlabel} are shown in Figure~\ref{Fig: all tempting exp dataset}.
\subsection{Algorithm details}
% We provide more details about the \method\ implementation in this section.

\textbf{OASIS algorithm training diagram}
In this work, we use the cosine $\beta$ schedule~\cite{nichol2021improved} to calculate $\beta_t, t=1, ..., K$. Then we let $\alpha_t = 1 - \beta_t$, and $\bar{\alpha}_t=\prod_{i=1}^t \alpha_i$ and denote the state dimension as $m$. With these notations, we show the training process of the OASIS data generator for one epoch in Alg.~\ref{alg:OASIS-training}.
\begin{algorithm}
\caption{\method\ (training)}
\label{alg:OASIS-training}
\begin{algorithmic}[1]
\STATE \textbf{Input:} Original Dataset $\mathcal{D}$, predefined $\beta_t$ and $\bar{\alpha}_t$, diffusion core $\epsilon_\theta\left(\boldsymbol{x}_k, \boldsymbol{y}, k\right)$, learning rate $lr$, loss function $L(\cdot, \cdot)$.
\FOR{each sub-trajectory $\tau_i \in \mathcal{D}$}
    \STATE Extract the states from $\tau_i$: $\{(s_0, s_1, \ldots, s_T)\}$ \# $[T, m]$;
    \STATE Get the return conditions $\boldsymbol{y} = [C, R]$ associated with these sub-trajectories;
    \STATE With probability $p=0.25$ to mask the condition information as: $\boldsymbol{y} \leftarrow \varnothing$ 
    \STATE Get Gaussian Noise $noise = \Ncal(\boldsymbol{0}, \boldsymbol{I})$ \# $[T, m]$;
    \STATE Randomly sample time $t \in [0, ..., K-1]$;
    \STATE Calculate the forward sampling state $\boldsymbol{x}_{noise} = \bar{\alpha}_t * \tau_i + (1-\bar{\alpha}_t) * noise$;
    \STATE Apply initial state condition $\boldsymbol{x}_{noise}[0] \leftarrow s_0$;
    \STATE Reconstruct noisy sub trajectory $\boldsymbol{x}_{recon} = \epsilon_\theta\left(\boldsymbol{x}_{noise}, \boldsymbol{y}, k\right)$;
    \STATE Minimize the reconstruction loss $\theta \leftarrow \theta - lr * \nabla_{\theta} L(\boldsymbol{x}_{noise}, \boldsymbol{x}_{recon})$;
\ENDFOR
\STATE \textbf{Output:} Updated diffusion core $\epsilon_\theta\left(\boldsymbol{x}_k, \boldsymbol{y}, k\right)$;
\end{algorithmic}
\end{algorithm}

\begin{figure}[ht]
    \centering
    % \vspace{-8pt} 
    \includegraphics[width=0.8\linewidth]{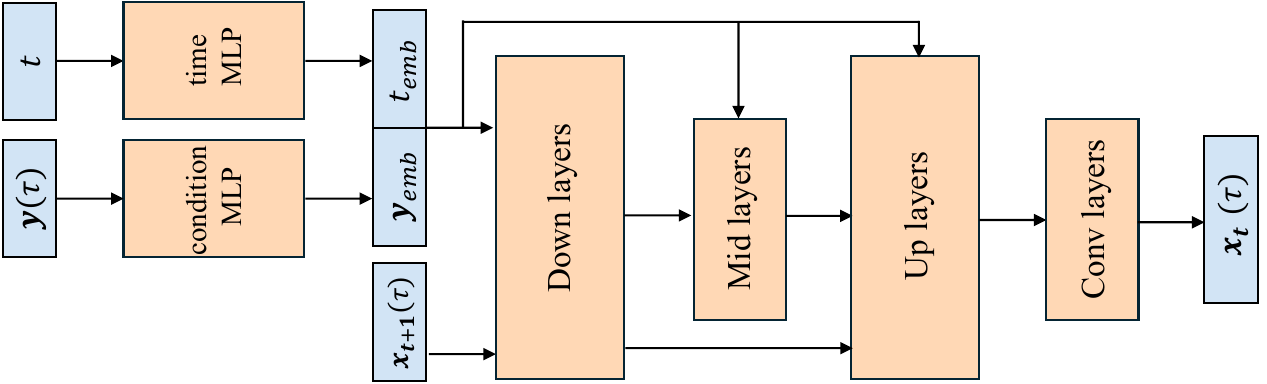}
    \caption{\texttt{U-Net} structure. }
    \label{Fig: layer}
\end{figure}

\textbf{Network and hyperparameter details.} For the dynamics model $\hat{p}$, we utilize a MLP. For the denoising core, we utilize a \texttt{U-net}, which has also been used in previous works~\cite{liang2023adaptdiffuser, ajay2022conditional}. The \texttt{U-net} is visualized in Figure~\ref{Fig: layer}. The hyperparameters for our method are summarized in Table~\ref{tab: hyperparameter}. More details are available in the code.

\textbf{Baseline details.} For the baseline methods \texttt{BC}, \texttt{BCQ-Lag}, \texttt{BEAR-Lag}, \texttt{COptiDICE}, \texttt{CPQ}, and \texttt{CDT}, we adopt the code base provided in the benchmark~\cite{liu2023datasets}. For the \texttt{FISOR} method, we use the code provided by the authors~\cite{zheng2024safe}.

\textbf{Computing resources.} The experiments are run on a server with 2$\times$AMD EPYC 7542 32-Core Processor CPU, 2$\times$NVIDIA RTX A6000 graphics, and $252$ GB memory. For one single experiment, \method\ takes about $4$ hours with $200,000$ steps to train the data generator. It takes about $1.5$ hours to train a \texttt{BCQ-Lag} agent on this generated dataset for $200,000$ steps.
\vspace{-7pt}
\begin{table}[htbp]
  \centering
  \caption{Hyperparameters}
    \begin{tabular}{cc}
    \toprule
    Hyperparameters & Value \\
    \midrule
    L (length of subsequence) & 32 \\
    K (denoising timestep) & 20 \\
    Batch size & 256 \\
    Learning rate & 3.0e-5 \\
    $w_\alpha$ & 2.0 \\
    \bottomrule
    \end{tabular}%
  \label{tab: hyperparameter}%
\end{table}%

\end{document}

%% file: math_commands.tex
\usepackage{amsmath,amsfonts,bm}
\usepackage{xspace}

\def\eqref#1{equation~\ref{#1}}
\def\1{\bm{1}}

\mathchardef\mhyphen="2D

\DeclareMathAlphabet{\mathsfit}{\encodingdefault}{\sfdefault}{m}{sl}
\SetMathAlphabet{\mathsfit}{bold}{\encodingdefault}{\sfdefault}{bx}{n}

\def\Acal{{\mathcal{A}}}

\def\Dcal{{\mathcal{D}}}

\def\Mcal{{\mathcal{M}}}
\def\Ncal{{\mathcal{N}}}
\def\Ocal{{\mathcal{O}}}
\def\Pcal{{\mathcal{P}}}

\def\Scal{{\mathcal{S}}}

\newcommand{\E}{\mathbb{E}}

\newcommand{\KL}{D_{\mathrm{KL}}}
\newcommand{\TV}{D_{\mathrm{TV}}}

\def\fs{{\mathbf{f}}}